\documentclass[journal]{IEEEtran}
\usepackage{amsmath,amssymb,amsfonts}
\usepackage{graphicx,booktabs,multirow,array}
\usepackage[table]{xcolor}
\usepackage{pifont,textcomp,url,cite}
\usepackage[caption=false,font=footnotesize]{subfig}
\usepackage{algorithmic}
\usepackage{placeins,capt-of}
\usepackage{needspace,flushend}
\usepackage{hyperref}
\AtBeginDocument{\DeclareFontShape{OT1}{ptm}{m}{scit}{<->ssub*ptm/m/it}{}}
\hypersetup{hidelinks}
\newcommand{\cmark}{\ding{51}}
\newcommand{\xmark}{\ding{55}}
\newcommand{\imp}[1]{\textcolor{green!50!black}{(+#1)}}
\newcounter{algorithm}
\newenvironment{nonfloatalgorithm}[1]{\refstepcounter{algorithm}\par\medskip\noindent\rule{\linewidth}{0.6pt}\par\vspace{2pt}\noindent\textbf{Algorithm \thealgorithm}\enspace #1\par\vspace{2pt}\noindent\rule{\linewidth}{0.4pt}\par\footnotesize}{\par\vspace{2pt}\noindent\rule{\linewidth}{0.6pt}\par\medskip}

\makeatletter
\def\ftype@dualtopoalgorithm{4}
\newenvironment{dualtopoalgorithmfloat}{\@float{dualtopoalgorithm}[!t]}{\end@float}
\makeatother
\let\dualtopobibliography\thebibliography
\renewcommand{\thebibliography}[1]{\vspace{-6pt}\dualtopobibliography{#1}\setlength{\baselineskip}{8.45pt}}
\author{Yiming~Yang, Hongbin~Lin, Yueru~Luo, Suzhong~Fu,\\
Chao~Zheng, Xinrui~Yan, Kun~Tang, Chen~Lv,~\IEEEmembership{Senior~Member,~IEEE},\\
Shuguang~Cui,~\IEEEmembership{Fellow,~IEEE}, and Zhen~Li%
\thanks{\textit{Corresponding author: Zhen Li (e-mail: \nolinkurl{lizhen@cuhk.edu.cn}).}}
\thanks{Yiming Yang, Hongbin Lin, Yueru Luo, and Suzhong Fu are with the Shenzhen Future Network of Intelligence Institute and Guangdong Provincial Key Laboratory of Future Networks of Intelligence, The Chinese University of Hong Kong (Shenzhen), Shenzhen 518172, China, and also with the School of Science and Engineering, The Chinese University of Hong Kong (Shenzhen), Shenzhen 518172, China (e-mail: \nolinkurl{yimingyang@link.cuhk.edu.cn}; \nolinkurl{223010185@link.cuhk.edu.cn}; \nolinkurl{222010057@link.cuhk.edu.cn}; \nolinkurl{224010165@link.cuhk.edu.cn}).}
\thanks{Chao Zheng, Xinrui Yan, and Kun Tang are with Tencent T-Lab, China (e-mail: \nolinkurl{chrisczheng@tencent.com}; \nolinkurl{xinruiyan@tencent.com}; \nolinkurl{kunntang@tencent.com}).}
\thanks{Chen Lv is with the School of Mechanical and Aerospace Engineering, Nanyang Technological University, Singapore 639798 (e-mail: \nolinkurl{lyuchen@ntu.edu.sg}).}
\thanks{Shuguang Cui and Zhen Li are with the School of Science and Engineering, The Chinese University of Hong Kong (Shenzhen), Shenzhen 518172, China, and also with the Shenzhen Future Network of Intelligence Institute and Guangdong Provincial Key Laboratory of Future Networks of Intelligence, The Chinese University of Hong Kong (Shenzhen), Shenzhen 518172, China (e-mail: \nolinkurl{shuguangcui@cuhk.edu.cn}; \nolinkurl{lizhen@cuhk.edu.cn}).}}

\title{DualTopo: Spatio-Temporal Lane Graph Generation with Dual Latent World Models and Parallel Query Supervision}
\begin{document}
\maketitle

\begin{abstract}
Lane topology reasoning constructs a spatio-temporal road graph $\mathcal{G}=(\mathcal{V},\mathcal{E})$ in bird's-eye view (BEV), providing structured road representations for planning-oriented autonomous driving. Stream-based methods propagate temporal cues at both the query and BEV levels, yet suffer from two critical limitations: \textbf{(1)} initialized queries receive insufficient supervision, as temporal queries dominate Hungarian matching and leave them undertrained; and \textbf{(2)} temporal propagation is weakened by BEV boundary information loss during feature warping. Since initialized queries serve as propagation anchors, particularly at the first frame, these deficiencies compound across frames and degrade graph generation quality. We propose \textbf{DualTopo}, a spatio-temporal lane graph generation framework with \textit{dual} latent world models and \textit{dual} parallel decoder pathways. The dual-pathway decoder supervises temporal and initialized queries concurrently, ensuring initialized queries are well-trained and serve as reliable anchors. Dual latent world models conditioned on ego-motion propagate latent state representations from past observations to the present, substantially improving spatio-temporal graph generation. Extensive experiments on OpenLane-V2 demonstrate leading lane detection accuracy and competitive topology graph prediction. Code is available at \url{https://github.com/YimingYang23/DualTopo}.
\end{abstract}

\begin{figure}[t]
   \centering
   \includegraphics[width=\linewidth]{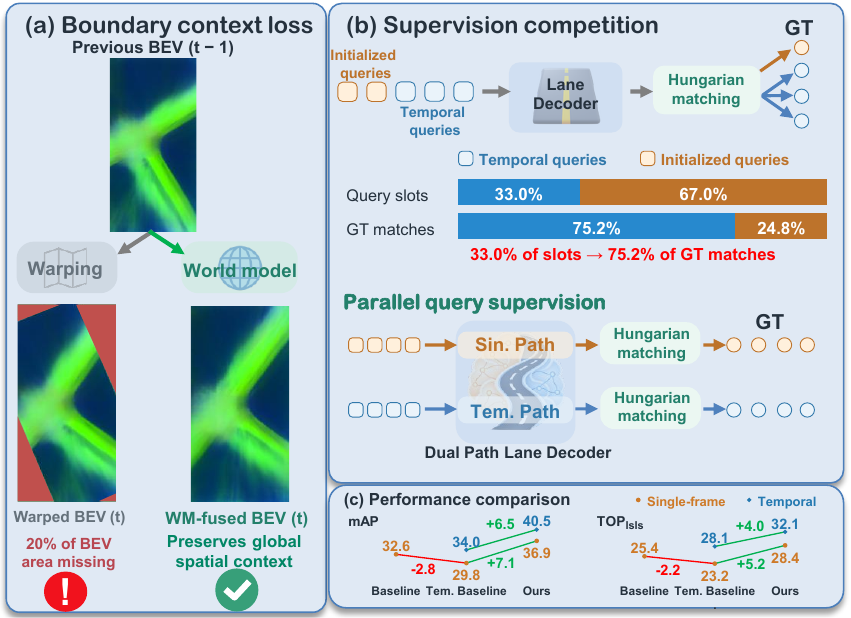}
\caption{Motivation for DualTopo. (a) Warping loses boundary context as ego-motion exposes regions outside the historical BEV grid (20\% in this example). The world model predicts BEV features across the current grid, which are fused with current observations. (b) Without parallel query supervision (PQS), temporal queries occupy 33.0\% of query slots but receive 75.2\% of ground-truth (GT) matches. PQS uses two decoder pathways with shared weights and independent Hungarian matching. (c) Conventional temporal propagation improves temporal outputs at the expense of single-frame accuracy. DualTopo yields substantial gains in both outputs, improving single-frame and temporal mAP by 7.1 and 6.5 points, respectively, over conventional temporal training.}
   \label{fig:motivation}
\end{figure}

\section{Introduction}
Lane topology reasoning converts surround-view images into a road graph containing lane geometry and connectivity \cite{wang2024openlane,li2024lanesegnet}. In bird's-eye view (BEV), graph nodes represent lanes and directed edges encode their adjacency. This structured representation supports planning-oriented autonomous driving \cite{hu2023planning}. Reliable graph generation requires accurate spatial predictions and consistent use of observations across time.

Recent methods strengthen graph construction \cite{lv2025t2sg}, endpoint alignment \cite{fu2024topologic}, and lane attribute prediction \cite{li2024lanesegnet}, but do not exploit temporal information. In practical deployment, camera observations arrive as a continuous stream, providing temporal cues for more accurate and consistent graph generation. To exploit these cues, stream-based models extend spatial representations by propagating historical queries and BEV features \cite{yuan2024streammapnet}. However, conventional temporal propagation introduces two issues that can degrade prediction performance, as illustrated in Figure~\ref{fig:motivation}. \textbf{Boundary context loss during BEV propagation.} Coordinate warping aligns historical BEV features with the current ego frame. Vehicle motion exposes boundary regions outside the support of the historical grid, where coordinate resampling cannot supply features. These regions occupy 20\% of the BEV area in Figure~\ref{fig:motivation}(a). \textbf{Competition for supervision within a shared decoder.} Joint Hungarian matching makes initialized and temporal queries compete for the same ground-truth (GT) assignments. High-confidence temporal predictions can dominate this competition, concentrating matched supervision on temporal queries. Under conventional stream-based learning \cite{yuan2024streammapnet}, temporal queries occupy 33.0\% of query slots but receive 75.2\% of GT matches. Initialized queries must support current-frame prediction, including the first frame, yet receive a much smaller share of matched supervision. In the single-frame ablation, conventional temporal training reduces mAP from 32.6 to 29.8 and TOP$_{lsls}$ from 25.4 to 23.2 (Figure~\ref{fig:motivation}(c)).

To address these challenges, we propose \textbf{DualTopo}, which combines \textbf{dual latent world models} with \textbf{parallel query supervision (PQS)} for spatio-temporal lane graph generation. A query world model (QWM) and BEV world model (BWM) predict current-frame latent states from historical representations and relative ego-motion. PQS decodes initialized and temporal queries through two pathways with shared parameters and independent Hungarian matching. Both pathways receive direct GT supervision, strengthening initialized queries that also serve as temporal propagation anchors. As illustrated in Figure~\ref{fig:motivation}, the world models predict current-frame latent states from history and ego-motion, while PQS resolves the supervision imbalance, yielding considerable improvements in both single-frame and temporal lane graph prediction.

\Needspace{5\baselineskip}
\noindent\textbf{Contributions:}
\begin{itemize}
\item We identify boundary context loss and imbalanced query supervision in streaming lane graph generation, and introduce \textbf{parallel query supervision} through two decoder pathways with shared weights and independent Hungarian matching to directly supervise initialized and temporal queries.
\item We design \textbf{dual latent world models} conditioned on ego-motion to propagate BEV features, query representations, and directed graph context across frames.
\item Experiments on OpenLane-V2 \cite{wang2024openlane} demonstrate leading temporal lane segment detection and topology scores among the compared methods. Adaptations to multiple baselines and ablation studies support the effectiveness of the propagation and supervision design.
\end{itemize}

\begin{figure*}[t]
   \centering
   \includegraphics[width=0.80\textwidth]{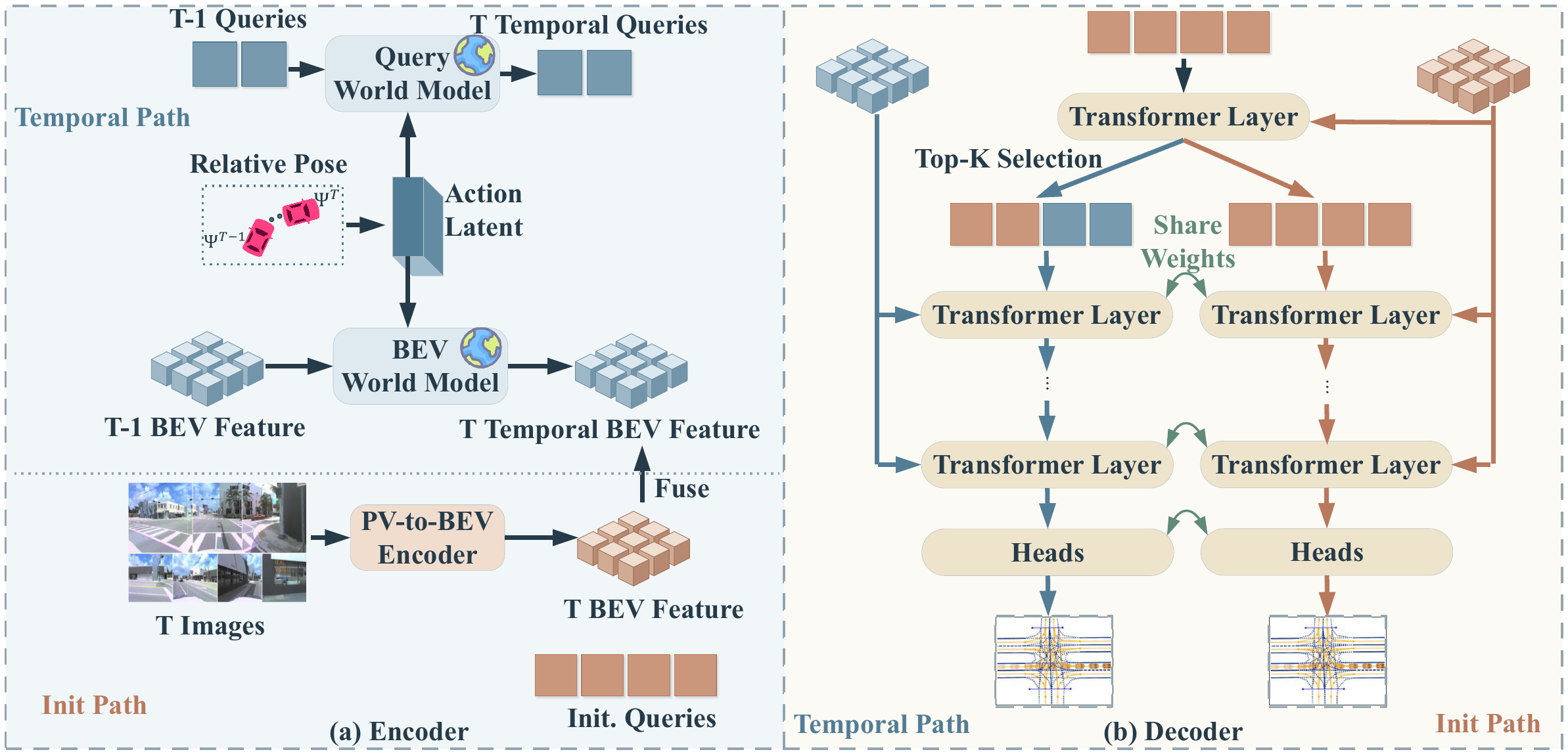}
\caption{DualTopo framework. (a) Encoder and world models. Relative ego-motion is encoded into the action latent $\Psi$. Conditioned on $\Psi$, the QWM and BWM predict temporal queries $\tilde{\mathbf{Q}}^T$ and temporal BEV features $\tilde{\mathbf{F}}^T$ from historical states. The predicted BEV features are fused with current-frame observations. (b) Parallel decoding. The temporal and init pathways share Transformer layers and prediction heads. Independent Hungarian matching supplies direct supervision to both pathways. $T$ denotes the current frame.}
   \label{fig:framework}
\end{figure*}

\section{Related Work}
\subsection{HD Map and Lane Topology Reasoning}
End-to-end HD map learning represents road elements as vectors predicted from sensor observations. VectorMapNet \cite{liu2023vectormapnet} uses polylines, while MapTR \cite{liao2022maptr,liao2023maptrv2} models permutation-equivalent lane point sets. Bai et al. \cite{11410107} further investigate dynamic query management and instance-level representation consistency for online vectorized HD map construction. Temporal extensions include stream-based propagation in StreamMapNet \cite{yuan2024streammapnet}, query denoising in SQD-MapNet \cite{wang2024stream}, and local-to-global fusion in InteractionMap \cite{wu2025interactionmap}.

Lane topology reasoning jointly predicts road geometry and the relationships between road elements. TopoNet \cite{li2023graph} models lane-to-lane and lane-to-traffic-element graphs. TopoLogic \cite{fu2024topologic}, TopoPoint \cite{fu2025topopoint}, and TopoFG \cite{xu2025fine} use endpoint information to improve connectivity prediction. TopoMLP \cite{wu2024topomlp} explicitly encodes coordinates. TopoFormer \cite{lv2025t2sg} uses geometry-guided and counterfactual self-attention, while RATopo \cite{li2025ratopo} addresses label redundancy through assignment-aware training. Beyond lane graphs, Shi et al. \cite{11561544} model object interactions in traffic scene graphs and exploit node and edge histories for traffic anomaly detection. 

TopoStreamer \cite{yang2025topostreamer} extends lane topology reasoning through temporal query propagation and attribute-aware denoising. It propagates BEV features by warping and jointly decodes temporal and initialized queries. DualTopo assigns temporal and current-frame inputs to separate pathways, combining independent supervision with ego-motion-conditioned latent propagation.

\subsection{World Model in Autonomous Driving}
\label{sec:world_model_rw}
Action-conditioned world models predict state evolution from observations and ego actions. Latent models perform this transition in compact feature spaces. BEVWorld \cite{zhang2024bevworld} models latent BEV sequences using diffusion, WoTE \cite{li2025end} predicts BEV features for trajectory scoring, and LAW \cite{li2024enhancinge2e} supports self-supervised end-to-end learning.

DualTopo uses latent state prediction for current-frame road graph generation. Known relative ego-motion conditions the transition $\mathbf{z}^{T-1} \xrightarrow{\mathcal{M}(\cdot,\mathbf{a})} \mathbf{z}^{T}$, whose output supports graph decoding at frame $T$. Separate world models operate on BEV features and lane queries, carrying spatial context and structured lane representations across frames.

\section{Method}
\subsection{Problem Formulation}
\textbf{Lane Graph Generation.} Given surround-view images $\mathcal{I}^T$ at frame $T$, we predict a directed road graph $\mathcal{G}^T=(\mathcal{V},\mathcal{E})$ in BEV, where $\mathcal{V}$ is the set of road-element nodes and $\mathcal{E}\subseteq\mathcal{V}\times\mathcal{V}$ is the set of directed topological edges. Lane nodes encode geometry, and the adjacency matrix $\mathbf{A}$ encodes lane-to-lane connectivity. We consider two graph representations. In a lane segment graph, each lane node comprises a centerline $\mathbf{L}^c=(\mathbf{P},\mathit{Class})$ and left and right boundaries, $\mathbf{L}^l=(\mathbf{P},\mathit{Type})$ and $\mathbf{L}^r=(\mathbf{P},\mathit{Type})$. Each $\mathbf{P}=\{(x_i,y_i,z_i)\}_{i=1}^M$ contains the ordered 3D points of the corresponding curve; $\mathit{Class}$ denotes the lane category, and $\mathit{Type}$ denotes boundary markings such as dashed or solid. A centerline--traffic element graph represents lanes by centerlines without boundary curves. Its node set additionally includes traffic elements, and its edge set includes both lane-to-lane and lane-to-traffic-element relations. DualTopo applies temporal modeling and parallel supervision to lane geometry and topology while retaining each backbone's traffic-element detection branch.

\subsection{Overview}
Figure~\ref{fig:framework} presents the overall architecture of DualTopo, which combines motion-conditioned latent propagation with parallel decoding. We use $T$ for the frame index and $i$ for the layer index in an $L$-layer decoder. The PV-to-BEV encoder \cite{li2022bevformer,he2016deep,lin2017feature} extracts current-frame features $\mathbf{F}^T \in \mathbb{R}^{H \times W \times C}$, where $H$ and $W$ are the longitudinal and lateral grid sizes, respectively, and $C$ is the feature dimension. A learnable embedding $\mathbf{Q}_0 \in \mathbb{R}^{N \times C}$ supplies $N$ initialized queries. Conditioned on the relative-pose latent $\Psi$, the QWM and BWM transform the previous-frame query output $\tilde{\mathbf{Q}}_L^{T-1}$ and fused BEV features $\tilde{\mathbf{F}}^{T-1}$ into predictions $\tilde{\mathbf{Q}}^T$ and $\tilde{\mathbf{F}}^T$. A gated recurrent unit (GRU) \cite{chung2014empirical} then fuses this prediction with the observed features $\mathbf{F}^T$, updating the temporal BEV state $\tilde{\mathbf{F}}^T$ in place.

After $i$ decoder layers, $\mathbf{Q}_i^T$ and $\tilde{\mathbf{Q}}_i^T$ denote the init and temporal query states, respectively. The first layer updates initialized queries using $\mathbf{F}^T$; subsequent layers process both pathways with shared weights and independent Hungarian matching. The temporal pathway provides the primary inference output. Its final queries $\tilde{\mathbf{Q}}_L^T$ and fused features $\tilde{\mathbf{F}}^T$ are cached for the next frame.

\subsection{Temporal Propagation via Latent World Models}
\textbf{Coordinate Warping and Boundary Context.} Coordinate warping aligns historical BEV features with the current ego frame through pose-based resampling \cite{yuan2024streammapnet}. Ego-motion exposes boundary regions outside the historical grid, where resampling cannot supply features. We therefore learn ego-motion-conditioned transitions for BEV features and lane queries.

\textbf{Action-Aware Latent Inputs.} The action latent $\Psi\in\mathbb{R}^{12}$ flattens the rotation and translation from the previous ego frame to the current frame. We concatenate $\Psi$ with the historical features and use multilayer perceptrons (MLPs) to form the motion-conditioned inputs:
\begin{equation}
	\label{eq:action_input}
    \begin{split}
	\hat{\mathbf{Q}}^{T-1} &= \textbf{MLP}([\tilde{\mathbf{Q}}_L^{T-1};\, \Psi])\\
    \hat{\mathbf{F}}^{T-1} &= \textbf{MLP}([\tilde{\mathbf{F}}^{T-1};\, \Psi]),
     \end{split}
\end{equation}
where $\Psi$ is replicated across query slots or BEV locations before channel concatenation.

\textbf{Latent State Prediction.} The world models propagate action-aware latent representations from $T\!-\!1$ to generate temporal features at $T$:
\begin{equation}
	\label{eq:wm_prediction}
    \begin{split}
	\tilde{\mathbf{Q}}^T &= \textbf{QWM}(\hat{\mathbf{Q}}^{T-1})\\
    \tilde{\mathbf{F}}^T &= \textbf{BWM}(\hat{\mathbf{F}}^{T-1})
     \end{split}
\end{equation}
The QWM consists of Transformer blocks with self-attention and feed-forward modules. Each BWM layer applies deformable self-attention, using $\hat{\mathbf{F}}^{T-1}$ as its queries, keys, and values. The BWM therefore predicts the current latent state from motion-conditioned historical features. 

\textbf{Directed Topology Propagation.} The query state also carries directed graph connectivity across frames. We cache the predicted adjacency matrix with the same ordered Top-$K$ queries. Its rows and columns remain aligned with the propagated queries, and invalid nodes and self-edges are masked. Let $\mathbf{A}^{T-1}$ denote their masked adjacency. Starting from the QWM prediction in Equation~\eqref{eq:wm_prediction}, we aggregate predecessor and successor context and update $\tilde{\mathbf{Q}}^T$ in place:
\begin{equation}
\label{eq:graph_latent_propagation}
\tilde{\mathbf{Q}}^T \leftarrow \tilde{\mathbf{Q}}^T
+\lambda\,\mathcal{F}_g\bigl(\tilde{\mathbf{Q}}^T,\mathbf{A}^{T-1}\bigr),
\end{equation}
where $\lambda$ controls the update strength, and $\mathcal{F}_g$ aggregates features from each lane's predecessors and successors. It applies separate learned affine projections and aggregates the projected features with row-normalized $(\mathbf{A}^{T-1})^{\top}$ and $\mathbf{A}^{T-1}$, respectively. The denominator is floored at $10^{-6}$ during row normalization. An MLP fuses the concatenated query, predecessor aggregate, and successor aggregate. Each valid lane's update is gated by its largest incoming or outgoing edge probability; invalid lanes have zero gates. The updated $\tilde{\mathbf{Q}}^T$ supplies the temporal decoder input. We cache raw relationship predictions, detach history between frames, and reset memory at scene boundaries.

\textbf{Latent Supervision.} Before fusion, the BWM prediction $\tilde{\mathbf{F}}^T$ is supervised by the current-frame BEV features $\mathbf{F}^T$:
\begin{equation}
	\label{eq:bev_loss}
	\mathcal{L}_{\text{bev}} = \operatorname{MSE}\!\left(\tilde{\mathbf{F}}^T,\,\mathbf{F}^T\right)
\end{equation}
The QWM receives transformation supervision \cite{yang2025topostreamer} for lane geometry, attributes, and masks. We suppress loss weights below and specify them in the supplementary material:
\begin{equation}
	\label{eq:query_loss}
    \begin{split}
	\mathcal{L}_{\text{coord}} &= \mathcal{L}_{L1}(\tilde{\mathbf{L}}^T,\, \mathbf{L}^T)\\
	\mathcal{L}_{\text{attr}} &= \mathcal{L}_{\text{Focal}}(\tilde{\mathit{Class}}^T, \mathit{Class}^T) + \mathcal{L}_{\text{CE}}(\tilde{\mathit{Type}}^T, \mathit{Type}^T)\\
	\mathcal{L}_{\text{mask}} &= \mathcal{L}_{\text{CE}}(\tilde{\mathbf{M}}^T, \mathbf{M}^T) + \mathcal{L}_{\text{Dice}}(\tilde{\mathbf{M}}^T, \mathbf{M}^T)\\
	\mathcal{L}_{\text{query}} &= \mathcal{L}_{\text{coord}} + \mathcal{L}_{\text{attr}} + \mathcal{L}_{\text{mask}},
    \end{split}
\end{equation}
where $\tilde{\mathbf{L}}^T$, $\tilde{\mathit{Class}}^T$, $\tilde{\mathit{Type}}^T$, and $\tilde{\mathbf{M}}^T$ are coordinates, lane classes, boundary types, and masks predicted from $\tilde{\mathbf{Q}}^T$. Their targets, $\mathbf{L}^T$, $\mathit{Class}^T$, $\mathit{Type}^T$, and $\mathbf{M}^T$, are ground-truth annotations transformed from frame $T\!-\!1$ to $T$. The combined latent objective is:
\begin{equation}
	\label{eq:latent_loss}
	\mathcal{L}_{\text{latent}} = \mathcal{L}_{\text{bev}} + \mathcal{L}_{\text{query}}
\end{equation}
BEV alignment trains the BWM to predict scene features under ego-motion. Transformation supervision trains the QWM to produce geometrically and semantically consistent temporal queries for current-frame graph generation.

\begin{table*}[t]
    \centering
\caption{Lane segment graph generation on OpenLane-V2 SubsetA.}\label{tab:comparison_lane_segment}

    \setlength{\tabcolsep}{2.5pt}
    \small
    \begin{tabular}{l|c|c|c|c|cccc}
    \toprule
    Method & Venue & Backbone & Epoch & Temporal & mAP & AP$_{ls}$ & AP$_{ped}$ & TOP$_{lsls}$ \\ \midrule
          MapTR~\cite{liao2022maptr} & ICLR'23 & ResNet-50 & 24 & \xmark & 27.0 & 25.9 & 28.1 & - \\
       MapTRv2~\cite{liao2023maptrv2} & IJCV'25 & ResNet-50 & 24 & \xmark & 28.5 & 26.6 & 30.4 & - \\
      TopoNet~\cite{li2023graph} & ArXiv'23 & ResNet-50 & 24 & \xmark & 23.0 & 23.9 & 22.0 & - \\
       LaneSegNet~\cite{li2024lanesegnet} & ICLR'24 & ResNet-50 & 24 & \xmark & 32.6 & 32.3 & 32.9 & 25.4 \\
          TopoLogic~\cite{fu2024topologic} & NeurIPS'24 & ResNet-50 & 24 & \xmark & 33.2 & 33.0 & 33.4 & 30.8 \\
          Topo2Seq~\cite{yang2025topo2seq} & AAAI'25 & ResNet-50 & 24 & \xmark & 33.6 & 33.7 & 33.5 & 26.9 \\ \midrule
        \textbf{DualTopo} & - & ResNet-50 & 24 & \xmark & 36.9 & 35.2 & 38.7 & 28.4 \\ \midrule
            StreamMapNet~\cite{yuan2024streammapnet} & WACV'24 & ResNet-50 & 24 & \cmark & 20.3 & 22.1 & 18.6 & 13.2 \\
        SQD-MapNet~\cite{wang2024stream} & ECCV'24 & ResNet-50 & 24 & \cmark & 26.0 & 27.1 & 24.9 & 16.6 \\
        TopoStreamer~\cite{yang2025topostreamer} & ArXiv'25 & ResNet-50 & 24 & \cmark & 36.6 & 35.0 & 38.1 & 28.5 \\ \midrule
       \textbf{DualTopo} & - & ResNet-50 & 24 & \cmark & \textbf{40.5} & \textbf{39.5} & \textbf{41.4} & \textbf{32.1} \\
\midrule
\multicolumn{5}{c|}{\textit{Improv.}} & \textcolor{green!50!black}{(+3.9)} & \textcolor{green!50!black}{(+4.5)} & \textcolor{green!50!black}{(+3.3)} & \textcolor{green!50!black}{(+1.3)} \\
       \bottomrule
    \end{tabular}
    
\end{table*}

\begin{table*}[t]
\centering
\caption{Centerline and traffic element graph generation on OpenLane-V2 SubsetA and SubsetB under V2.1.0 evaluation. $^{\S}$: our TopoFormer reimplementation.}
\label{tab:centerline_full_comparison}
\setlength{\tabcolsep}{2.5pt}
\footnotesize
\renewcommand{\arraystretch}{1.05}
\begin{tabular}{c|l|c|c|c|c|ccccc}
\toprule
Subset & Method & Venue & Backbone & Epoch & Temporal & DET$_l\uparrow$ & DET$_t\uparrow$ & TOP$_{ll}\uparrow$ & TOP$_{lt}\uparrow$ & OLS$\uparrow$ \\ \midrule
% SubsetA block
 & VectorMapNet~\cite{liu2023vectormapnet} & ICML2023 & ResNet-50 & 24 & \xmark & 11.1 & 41.7 & 2.7 & 9.2 & 24.9 \\
 & MapTR~\cite{liao2022maptr} & ICLR2023 & ResNet-50 & 24 & \xmark & 17.7 & 43.5 & 5.9 & 15.1 & 31.0 \\
 & TopoNet~\cite{li2023graph} & ArXiv2023 & ResNet-50 & 24 & \xmark & 28.6 & 48.6 & 10.9 & 23.8 & 39.8 \\
 & TopoMLP~\cite{wu2024topomlp} & ICLR2024 & ResNet-50 & 24 & \xmark & 28.5 & 50.5 & 21.7 & 27.3 & 44.5 \\
 & TopoFormer$^{\S}$~\cite{lv2025t2sg} & CVPR2025 & ResNet-50 & 24 & \xmark & 31.5 & 47.8 & 22.6 & 26.8 & 44.7 \\
 & CoPo~\cite{luo2026copo} & KDD2026 & ResNet-50 & 24 & \xmark & 33.8 & 50.9 & 29.2 & 32.2 & 48.9 \\
 & UniTopo~\cite{11506407} & TCSVT2026 & ResNet-50 & 24 & \xmark & 29.8 & 49.3 & 29.6 & 31.2 & 47.3 \\
\cmidrule(lr){2-11}

 & TopoLogic~\cite{fu2024topologic} & NeurIPS2024 & ResNet-50 & 24 & \xmark & 29.9 & 47.2 & 23.9 & 25.4 & 44.1 \\
  \rowcolor{gray!12}
\cellcolor{white} & \textbf{Ours(TopoLogic)} & -- & ResNet-50 & 24 & \cmark & 35.5\,\textcolor{green!50!black}{(+5.6)} & 47.2\,\textcolor{black}{(0.0)} & 31.9\,\textcolor{green!50!black}{(+8.0)} & 29.8\,\textcolor{green!50!black}{(+4.4)} & 48.4\,\textcolor{green!50!black}{(+4.3)} \\
\cmidrule(lr){2-11}
 & RATopo~\cite{li2025ratopo} & ACM MM2025 & ResNet-50 & 24 & \xmark & 31.8 & 49.4 & 32.2 & 33.9 & 49.0 \\
   \rowcolor{gray!12}
\cellcolor{white}    & \textbf{Ours(RATopo)} & -- & ResNet-50 & 24 & \cmark & 33.5\,\textcolor{green!50!black}{(+1.7)} & 49.8\,\textcolor{green!50!black}{(+0.4)} & \textbf{34.7}\,\textcolor{green!50!black}{(+2.5)} & \textbf{34.1}\,\textcolor{green!50!black}{(+0.2)} & 50.1\,\textcolor{green!50!black}{(+1.1)} \\
  \cmidrule(lr){2-11}
 & TopoFG~\cite{xu2025fine} & AAAI2026 & ResNet-50 & 24 & \xmark & 33.8 & 47.2 & 30.8 & 30.9 & 48.0 \\
   \rowcolor{gray!12}
\cellcolor{white}    & \textbf{Ours(TopoFG)} & -- & ResNet-50 & 24 & \cmark & \textbf{36.7}\,\textcolor{green!50!black}{(+2.9)} & 49.0\,\textcolor{green!50!black}{(+1.8)} & 31.6\,\textcolor{green!50!black}{(+0.8)} & 31.2\,\textcolor{green!50!black}{(+0.3)} & 49.4\,\textcolor{green!50!black}{(+1.4)} \\
  \cmidrule(lr){2-11}
 & TopoPoint~\cite{fu2025topopoint} & NeurIPS2025 & ResNet-50 & 24 & \xmark & 29.4 & 55.0 & 28.6 & 30.6 & 48.3 \\
  \rowcolor{gray!12}
\cellcolor{white}\multirow{-15}{*}{setA}  & \textbf{Ours(TopoPoint)} & -- & ResNet-50 & 24 & \cmark & 32.4\,\textcolor{green!50!black}{(+3.0)} & \textbf{55.1}\,\textcolor{green!50!black}{(+0.1)} & 33.7\,\textcolor{green!50!black}{(+5.1)} & 32.7\,\textcolor{green!50!black}{(+2.1)} & \textbf{50.7}\,\textcolor{green!50!black}{(+2.4)} \\
\midrule
% SubsetB block

 & TopoNet~\cite{li2023graph} & ArXiv2023 & ResNet-50 & 24 & \xmark & 24.3 & 55.0 & 6.7 & 16.7 & 36.5 \\
 & TopoMLP~\cite{wu2024topomlp} & ICLR2024 & ResNet-50 & 24 & \xmark & 26.0 & 58.2 & 21.0 & 19.8 & 43.6 \\
 & TopoFormer$^{\S}$~\cite{lv2025t2sg} & CVPR2025 & ResNet-50 & 24 & \xmark & 29.2 & 55.1 & 20.9 & 19.7 & 43.6 \\
 & CoPo~\cite{luo2026copo} & KDD2026 & ResNet-50 & 24 & \xmark & 32.6 & 58.8 & 31.8 & 25.8 & 49.7 \\
 & UniTopo~\cite{11506407} & TCSVT2026 & ResNet-50 & 24 & \xmark & 29.4 & 58.6 & 28.3 & 23.4 & 47.4 \\
\cmidrule(lr){2-11}
 & TopoLogic~\cite{fu2024topologic} & NeurIPS2024 & ResNet-50 & 24 & \xmark & 25.9 & 54.7 & 21.6 & 17.9 & 42.3 \\
  \rowcolor{gray!12}
\cellcolor{white} & \textbf{Ours(TopoLogic)} & -- & ResNet-50 & 24 & \cmark & 33.4\,\textcolor{green!50!black}{(+7.5)} & 55.3\,\textcolor{green!50!black}{(+0.6)} & 31.9\,\textcolor{green!50!black}{(+10.3)} & 21.7\,\textcolor{green!50!black}{(+3.8)} & 48.0\,\textcolor{green!50!black}{(+5.7)} \\
\cmidrule(lr){2-11}
 & RATopo~\cite{li2025ratopo} & ACM MM2025 & ResNet-50 & 24 & \xmark & 33.1 & 54.7 & 37.3 & 27.4 & 50.3 \\
   \rowcolor{gray!12}
\cellcolor{white} & \textbf{Ours(RATopo)} & -- & ResNet-50 & 24 & \cmark & \textbf{41.5}\,\textcolor{green!50!black}{(+8.4)} & 56.7\,\textcolor{green!50!black}{(+2.0)} & \textbf{45.9}\,\textcolor{green!50!black}{(+8.6)} & \textbf{29.1}\,\textcolor{green!50!black}{(+1.7)} & \textbf{55.0}\,\textcolor{green!50!black}{(+4.7)} \\
  \cmidrule(lr){2-11}
 & TopoFG~\cite{xu2025fine} & AAAI2026 & ResNet-50 & 24 & \xmark & 30.0 & 53.0 & 27.2 & 21.7 & 45.4 \\
   \rowcolor{gray!12}
\cellcolor{white}    & \textbf{Ours(TopoFG)} & -- & ResNet-50 & 24 & \cmark & 35.5\,\textcolor{green!50!black}{(+5.5)} & 55.4\,\textcolor{green!50!black}{(+2.4)} & 32.8\,\textcolor{green!50!black}{(+5.6)} & 22.3\,\textcolor{green!50!black}{(+0.6)} & 48.8\,\textcolor{green!50!black}{(+3.4)} \\
  \cmidrule(lr){2-11}
 & TopoPoint~\cite{fu2025topopoint} & NeurIPS2025 & ResNet-50 & 24 & \xmark &31.2 &\textbf{60.2} &28.3 &27.1 &49.2 \\
  \rowcolor{gray!12}
\cellcolor{white}\multirow{-13}{*}{setB} & \textbf{Ours(TopoPoint)} & -- & ResNet-50 & 24 & \cmark & 34.3\,\textcolor{green!50!black}{(+3.1)} & 60.1\,\textcolor{red!80!black}{(-0.1)} & 37.3\,\textcolor{green!50!black}{(+9.0)} & 28.6\,\textcolor{green!50!black}{(+1.5)} & 52.2\,\textcolor{green!50!black}{(+3.0)} \\
\bottomrule
\end{tabular}

\end{table*}

\begin{table*}[t]
\centering
\begin{minipage}[t]{0.485\textwidth}\vspace{0pt}
\centering
\captionof{table}{Module combinations on SubsetA. The second row enables only Tem. and uses BEV warping. Gray: full model. Improvements: Tem. vs. temporal-only; Sin. vs. single-frame baseline.}\label{tab:Ablation_different_config}\label{tab:Ablation_modules}
\small\renewcommand{\arraystretch}{1.04}\setlength{\tabcolsep}{2.5pt}
\begin{tabular}{ccccccll}
\toprule
\multicolumn{4}{c}{Modules} & \multicolumn{2}{c}{Tem.} & \multicolumn{2}{c}{Sin.} \\ \cmidrule(lr){5-6}\cmidrule(lr){7-8}
Tem. & PQS & QWM & BWM & mAP & TOP$_{lsls}$ & mAP & TOP$_{lsls}$ \\ \midrule
\xmark & \xmark & \xmark & \xmark & -- & -- & 32.6 & 25.4 \\
\cmark & \xmark & \xmark & \xmark & 34.0 & 28.1 & \textcolor{red!80!black}{29.8\,$\downarrow$} & \textcolor{red!80!black}{23.2\,$\downarrow$} \\
\cmark & \cmark & \xmark & \xmark & 38.6 & 30.5 & 36.4 & 27.8 \\
\cmark & \cmark & \cmark & \xmark & 38.9 & 31.1 & 35.4 & 28.1 \\
\cmark & \cmark & \xmark & \cmark & 38.9 & 31.1 & 36.5 & 27.9 \\
\cmark & \xmark & \cmark & \cmark & 38.9 & 30.2 & 34.2 & 25.7 \\
\rowcolor{gray!8} \cmark & \cmark & \cmark & \cmark & \textbf{40.5} & \textbf{32.1} & \textbf{36.9} & \textbf{28.4} \\
\midrule
\multicolumn{4}{c}{\textit{Improv.}} & \textcolor{green!50!black}{(+6.5)} & \textcolor{green!50!black}{(+4.0)} & \textcolor{green!50!black}{(+4.3)} & \textcolor{green!50!black}{(+3.0)} \\
\bottomrule
\end{tabular}

\end{minipage}\hfill
\begin{minipage}[t]{0.485\textwidth}\vspace{0pt}
\centering
\captionof{table}{Generality on StreamMapNet \cite{yuan2024streammapnet} (nuScenes). $\spadesuit$: dual world models; $\clubsuit$: additionally using parallel query supervision.}\label{tab:streammapnet_generality}

    \setlength{\tabcolsep}{2.5pt}
    \small
    \begin{tabular}{c|l|cccc}
    \toprule
    Split & Method & AP$_{ped}\uparrow$ & AP$_{div}\uparrow$ & AP$_{bound}\uparrow$ & AP$\uparrow$ \\ \midrule
    \multirow{3}{*}{Old} & StreamMapNet & 61.7 & 66.3 & 62.1 & 63.4 \\
    & StreamMapNet$^\spadesuit$ & 64.5 & 67.4 & 65.9 & 65.9 \\
    & StreamMapNet$^\clubsuit$ & \textbf{65.4} & \textbf{70.2} & \textbf{67.2} & \textbf{67.6} \\ \midrule
    & \textit{Improv.} & \imp{3.7} & \imp{3.9} & \imp{5.1} & \imp{4.2} \\ \midrule
    \multirow{3}{*}{New} & StreamMapNet & 32.2 & 29.3 & 40.8 & 34.1 \\
    & StreamMapNet$^\spadesuit$ & 34.8 & 31.8 & 42.6 & 36.4 \\
    & StreamMapNet$^\clubsuit$ & \textbf{35.6} & \textbf{32.4} & \textbf{44.9} & \textbf{37.6} \\ \midrule
    & \textit{Improv.} & \imp{3.4} & \imp{3.1} & \imp{4.1} & \imp{3.5} \\
    \bottomrule
    \end{tabular}

\end{minipage}

\par\vspace{10pt}
\centering\captionsetup{position=top}
\caption{Training and model configuration ablations on SubsetA. Gray rows indicate default settings. Panel (b) reports temporal and single-frame outputs; the other panels report temporal outputs.}\label{tab:world_model_config}
\footnotesize\renewcommand{\arraystretch}{1.05}\setlength{\tabcolsep}{1.8pt}
\subfloat[Training schedule and action condition.\label{tab:two_stage}\label{tab:wm_condition}]{\begin{minipage}[t]{0.32\textwidth}\vspace{0pt}\centering\footnotesize
\begin{tabular}{lcccc}
\toprule
Setting & mAP & AP$_{ls}$ & AP$_{ped}$ & TOP$_{lsls}$ \\ \midrule
\multicolumn{5}{l}{\textit{Training schedule}} \\
Scratch & 37.7 & 37.9 & 37.6 & 30.2 \\
\rowcolor{gray!8} Two-stage & \textbf{40.5} & \textbf{39.5} & \textbf{41.4} & \textbf{32.1} \\
\textit{Improv.} & \textcolor{green!50!black}{(+2.8)} & \textcolor{green!50!black}{(+1.6)} & \textcolor{green!50!black}{(+3.8)} & \textcolor{green!50!black}{(+1.9)} \\ \midrule
\multicolumn{5}{l}{\textit{Action conditioning}} \\
None & 38.1 & 37.5 & 38.8 & 30.6 \\
\rowcolor{gray!8} Relative pose & \textbf{40.5} & \textbf{39.5} & \textbf{41.4} & \textbf{32.1} \\
\midrule
\textit{Improv.} & \textcolor{green!50!black}{(+2.4)} & \textcolor{green!50!black}{(+2.0)} & \textcolor{green!50!black}{(+2.6)} & \textcolor{green!50!black}{(+1.5)} \\
\bottomrule
\end{tabular}
\end{minipage}}
\hfill
\subfloat[Training strategies.\label{tab:Ablation_training}]{\begin{minipage}[t]{0.32\textwidth}\vspace{0pt}\centering\footnotesize
\begin{tabular}{lcccc}
\toprule
Method & \multicolumn{2}{c}{Tem.} & \multicolumn{2}{c}{Sin.} \\ & mAP & TOP$_{lsls}$ & mAP & TOP$_{lsls}$ \\ \midrule
Temporal & 34.0 & 28.1 & 29.8 & 23.2 \\
Random+WMs & 38.3 & 31.1 & 35.5 & 27.8 \\
\rowcolor{gray!8} DualPath+WMs & \textbf{40.5} & \textbf{32.1} & \textbf{36.9} & \textbf{28.4} \\
\midrule
\textit{Improv.} & \textcolor{green!50!black}{(+2.2)} & \textcolor{green!50!black}{(+1.0)} & \textcolor{green!50!black}{(+1.4)} & \textcolor{green!50!black}{(+0.6)} \\
\bottomrule
\end{tabular}
\end{minipage}}
\hfill
\subfloat[World model architecture.\label{tab:wm_type}]{\begin{minipage}[t]{0.32\textwidth}\vspace{0pt}\centering\footnotesize
\begin{tabular}{lcccc}
\toprule
Arch. & mAP & AP$_{ls}$ & AP$_{ped}$ & TOP$_{lsls}$ \\ \midrule
Linear & 38.2 & 36.5 & 39.9 & 30.0 \\
MLPs & 38.0 & 35.6 & 40.4 & 31.0 \\
\rowcolor{gray!8} Transformer & \textbf{40.5} & \textbf{39.5} & \textbf{41.4} & \textbf{32.1} \\
\midrule
\textit{Improv.} & \textcolor{green!50!black}{(+2.3)} & \textcolor{green!50!black}{(+3.0)} & \textcolor{green!50!black}{(+1.5)} & \textcolor{green!50!black}{(+2.1)} \\
\bottomrule
\end{tabular}
\end{minipage}}
\par\vspace{7pt}
\subfloat[BEV history length.\label{tab:memory_length}]{\begin{minipage}[t]{0.32\textwidth}\vspace{0pt}\centering\footnotesize
\begin{tabular}{ccccc}
\toprule
Length & mAP & AP$_{ls}$ & AP$_{ped}$ & TOP$_{lsls}$ \\ \midrule
\rowcolor{gray!8} 1 & 40.5 & 39.5 & 41.4 & 32.1 \\
2 & \textbf{40.8} & \textbf{39.7} & \textbf{41.9} & \textbf{32.4} \\
4 & 38.4 & 36.8 & 39.9 & 31.5 \\
8 & 40.0 & 38.8 & 41.1 & 31.5 \\
\bottomrule
\end{tabular}
\end{minipage}}
\hfill
\subfloat[World model depth.\label{tab:wm_number}]{\begin{minipage}[t]{0.32\textwidth}\vspace{0pt}\centering\footnotesize
\begin{tabular}{ccccc}
\toprule
Layers & \multicolumn{2}{c}{QWM} & \multicolumn{2}{c}{BWM} \\ & mAP & TOP$_{lsls}$ & mAP & TOP$_{lsls}$ \\ \midrule
\rowcolor{gray!8} 2 & 40.5 & 32.1 & 40.5 & 32.1\\
4 & \textbf{40.7} & \textbf{32.3} & 40.9 & 32.2 \\
6 & \textbf{40.7} & 31.8 & \textbf{41.1} & \textbf{32.4} \\
\bottomrule
\end{tabular}
\end{minipage}}
\hfill
\subfloat[Temporal query count.\label{tab:k_compare}]{\begin{minipage}[t]{0.32\textwidth}\vspace{0pt}\centering\footnotesize
\begin{tabular}{lcccc}
\toprule
Top-$K$ & mAP & AP$_{ls}$ & AP$_{ped}$ & TOP$_{lsls}$ \\ \midrule
20 (10\%) & 37.8 & 38.1 & 37.5 & 29.8 \\
\rowcolor{gray!8} 66 (33\%) & \textbf{40.5} & \textbf{39.5} & \textbf{41.4} & \textbf{32.1}\\
100 (50\%) & 39.4 & 38.0 & 40.8 & 30.7 \\
150 (75\%) & 37.8 & 36.9 & 38.7 & 28.6 \\
\bottomrule
\end{tabular}
\end{minipage}}

\end{table*}

\subsection{Dual-Pathway Decoder}\label{sec:dual_path}
\begin{figure}[!t]
    \centering
    \includegraphics[width=0.88\columnwidth]{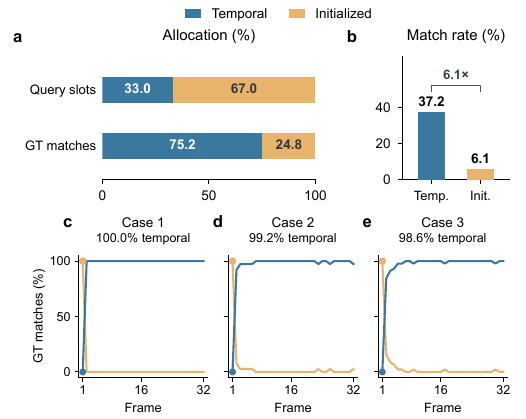}
    \caption{Matching supervision without PQS. (a) Query slots versus GT-match allocation. (b) Matches per available query. (c)--(e) Three complete 32-frame sequences. Initialized queries receive all GT matches at frame 1; temporal queries dominate afterward. }
    \label{fig:query_matching_supervision}

\end{figure}

Joint Hungarian matching makes temporal and initialized queries compete for the same GT assignments. Across non-first-frame records from the last decoder layer, temporal queries occupy 33.0\% of slots but receive 75.2\% of GT matches, yielding a 6.1-fold higher per-query match rate (Figure~\ref{fig:query_matching_supervision}(a)--(b)). In the extreme cases of Figure~\ref{fig:query_matching_supervision}(c)--(e), temporal queries receive 98.6--100.0\% of non-first-frame matches.

\begin{table}[!t]
    \centering
    \includegraphics[width=0.84\columnwidth]{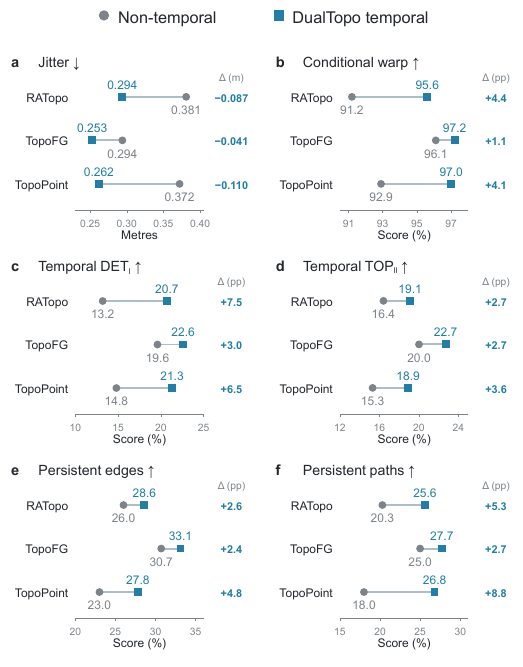}
    \captionof{figure}{Temporal consistency on SubsetA with three backbones. Grey circles denote non-temporal baseline outputs; blue squares denote DualTopo temporal outputs.}
    \label{fig:three_backbone_temporal}
\end{table}

\begin{figure*}[t]
    \centering
    \includegraphics[width=0.80\textwidth]{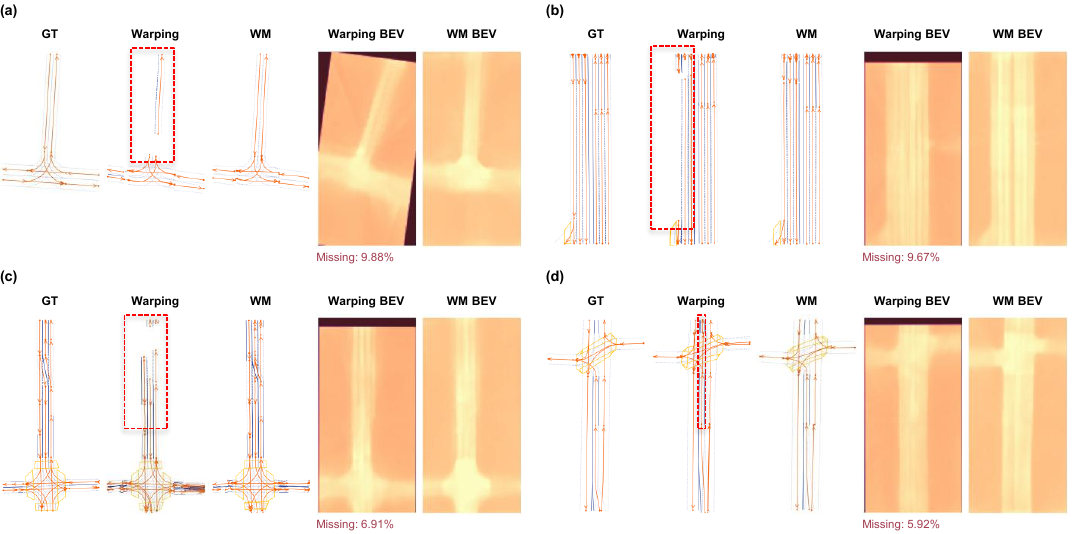}
    \caption{BEV propagation and lane predictions in four SubsetA frames. Each case shows GT, the trained Warping baseline, and DualTopo (WM), alongside the propagated BEV features. Red dashed boxes mark missing or incomplete lane segments in the Warping predictions.}
    \label{fig:wm_warping_bev}
\end{figure*}

\begin{figure*}[t]
    \centering
    \includegraphics[width=0.76\textwidth]{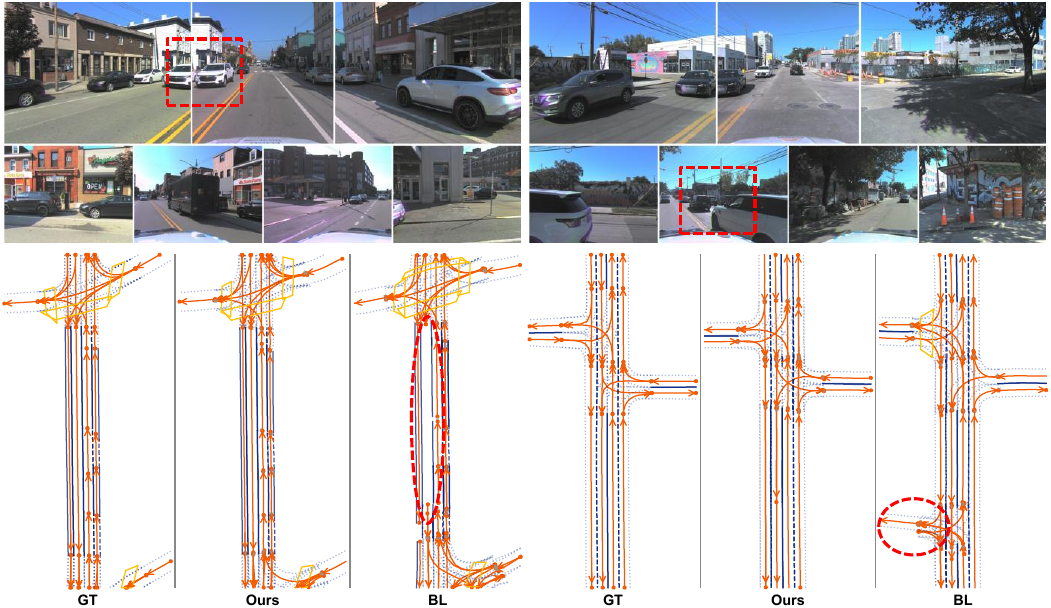}
    \caption{Qualitative comparison in two scenes with vehicle occlusion. Panels show multi-view images, ground truth (GT), DualTopo (Ours), and LaneSegNet (BL). Red dashed boxes mark occluding vehicles. Red dashed ellipses identify missed lane segments in BL (left) and a spurious topological branch absent from GT (right). DualTopo recovers more complete lane geometry and connectivity in these regions.}
    \label{fig:vis}
\par\vspace{8pt}
\centering
    \includegraphics[width=0.76\textwidth]{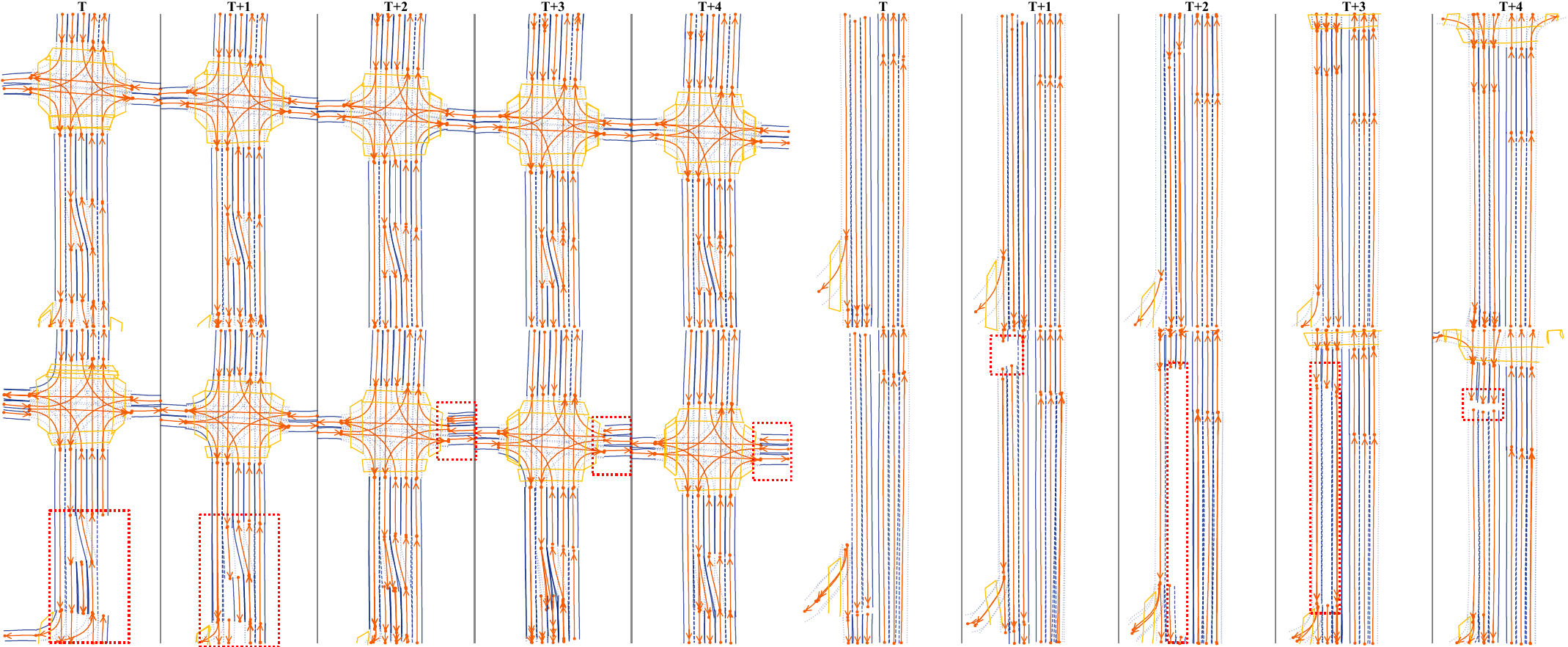}
    \caption{Five-frame lane segment predictions. Each sequence compares DualTopo (top) with LaneSegNet (bottom).}
    \label{fig:stream_compare}
\end{figure*}

We address this imbalance with PQS. Temporal and init pathways share decoder weights and prediction heads, while computing Hungarian assignments independently. Both query groups therefore receive direct GT supervision, strengthening current-frame prediction and the initialized queries used as propagation anchors.

As shown in Figure~\ref{fig:framework}(b), the first Transformer layer processes initialized queries with current BEV features:
\begin{equation}
	\label{eq:first_layer}
	\mathbf{Q}_1^T = \textbf{TransLayer}_0(\mathbf{Q}_0,\, \mathbf{F}^T),
\end{equation}
where $\textbf{TransLayer}_0$ contains self-attention, lane attention \cite{li2024lanesegnet}, and a feed-forward network. At non-first frames, we rank $\mathbf{Q}_1^T$ by classification confidence and replace its lowest-ranked $K$ queries with the top-$K$ temporal queries $\tilde{\mathbf{Q}}^T$. This forms the temporal input $\tilde{\mathbf{Q}}_1^T$. At the first frame, we set $\tilde{\mathbf{Q}}_1^T=\mathbf{Q}_1^T$. Before temporal decoding, we update the BEV state in place as $\tilde{\mathbf{F}}^T \leftarrow \text{GRU}(\tilde{\mathbf{F}}^T,\, \mathbf{F}^T)$. The remaining Transformer layers process both pathways with shared weights:
\begin{equation}
	\label{eq:temporal_branch}
	\tilde{\mathbf{Q}}_{i+1}^T = \textbf{TransLayer}_i(\tilde{\mathbf{Q}}_i^T,\, \tilde{\mathbf{F}}^T), \quad i = 1,\ldots,L-1
\end{equation}
\begin{equation}
	\label{eq:init_branch}
	\mathbf{Q}_{i+1}^T = \textbf{TransLayer}_i(\mathbf{Q}_i^T,\, \mathbf{F}^T), \quad i = 1,\ldots,L-1
\end{equation}
The temporal input contains $K$ temporal queries and $N\!-\!K$ updated initialized queries, with $N$ total slots. The temporal pathway attends to the updated temporal BEV features $\tilde{\mathbf{F}}^T$. The init pathway attends to the current-frame features $\mathbf{F}^T$. Shared heads predict lanes and topology from both $\tilde{\mathbf{Q}}_{i+1}^T$ and $\mathbf{Q}_{i+1}^T$. Each pathway computes its own Hungarian assignment during training.

Independent assignments provide direct supervision for initialized queries throughout training. The shared decoder and prediction heads also allow the init pathway to benefit from temporal learning, improving single-frame prediction from current-frame inputs. Both pathways retain their own query states and BEV inputs, supporting single-frame prediction and temporal decoding within one model.

\section{Training Loss}
\noindent\textbf{Lane Segment Loss.} We use the LaneSegNet loss \cite{li2024lanesegnet} to supervise lane geometry, semantic attributes, masks, and topology:
% Allow a column break before the first display without stranding the section title.
\begingroup\predisplaypenalty=0
\begin{equation}
	\label{eq:lanesegment}
\begin{aligned}
\mathcal{L}_{ls} &= \omega_1 \mathcal{L}_{\text{vec}} + \omega_2 \mathcal{L}_{\text{seg}} + \omega_3 \mathcal{L}_{\text{cls}} \\
&\quad + \omega_4 \mathcal{L}_{\text{type}} + \omega_5 \mathcal{L}_{\text{topo}}
\end{aligned}
\end{equation}
\endgroup
with $\omega_1\!=\!0.025$, $\omega_2\!=\!3.0$, $\omega_3\!=\!1.5$, $\omega_4\!=\!0.1$, $\omega_5\!=\!5.0$.
$\mathcal{L}_{\text{vec}}$ is the L1 loss between predicted and ground-truth vectorized lane points.
$\mathcal{L}_{\text{seg}} = \mathcal{L}_{\text{CE}} + \mathcal{L}_{\text{Dice}}$ supervises the BEV semantic mask.
$\mathcal{L}_{\text{cls}}$ and $\mathcal{L}_{\text{type}}$ are focal and cross-entropy losses for lane classification and boundary type prediction, respectively.
$\mathcal{L}_{\text{topo}}$ applies focal supervision to predicted graph edges.

\noindent\textbf{Hungarian Matching.} Both pathways use the same Hungarian assignment cost, combining geometry, mask, category, and boundary-type terms:
\begin{equation}
	\label{eq:matching}
    \mathcal{L}_{\text{match}} = \beta_1 \mathcal{L}_{\text{vec}} + \beta_2 \mathcal{L}_{\text{seg}} + \beta_3 \mathcal{L}_{\text{cls}} + \beta_4 \mathcal{L}_{\text{type}}
\end{equation}
with $\beta_1\!=\!0.025$, $\beta_2\!=\!3.0$, $\beta_3\!=\!1.5$, $\beta_4\!=\!0.01$. Each pathway computes its own matching assignment, so initialized and temporal queries receive direct supervision within their respective pathways.

\noindent\textbf{Joint Objective.} We optimize detection and latent prediction jointly through the temporal and init pathways:
\begin{equation}
	\label{eq:temporal_loss}
	\mathcal{L}_{\text{temporal}} = \alpha_1 \mathcal{L}_{ls} + \alpha_2 \mathcal{L}_{\text{latent}}, \qquad
	\mathcal{L}_{\text{init}} = \mathcal{L}_{ls},
\end{equation}
where $\mathcal{L}_{\text{latent}}$ is defined in Equation~\eqref{eq:latent_loss}. The full objective combines the two pathway losses:
\begin{equation}
	\label{eq:total_loss}
	\mathcal{L} = \mathcal{L}_{\text{temporal}} + \mathcal{L}_{\text{init}}
\end{equation}

\noindent\textbf{Centerline--Traffic Element Graph Generation.} For this task, we retain the loss functions and Hungarian matching costs of the corresponding baseline: TopoLogic \cite{fu2024topologic}, TopoPoint \cite{fu2025topopoint}, TopoFG \cite{xu2025fine}, and RATopo \cite{li2025ratopo}. The baseline task loss replaces $\mathcal{L}_{ls}$ in the joint objective, while the two pathways compute their Hungarian assignments independently.

\section{Experiments}

\subsection{Datasets and Metrics}
\label{sec:datasets}
\textbf{Datasets.} We evaluate on OpenLane-V2 \cite{wang2024openlane}, which contains two subsets of 1,000 surround-view scenes each. SubsetA is derived from Argoverse2 \cite{wilson2023argoverse} and provides seven camera views, while SubsetB is derived from nuScenes \cite{caesar2020nuscenes} and provides six. Both subsets include lane centerline and traffic-element annotations, together with lane-to-lane and lane-to-traffic-element topology. We evaluate lane segment graph generation on SubsetA, which additionally provides lane-boundary annotations; these annotations are unavailable in SubsetB. Centerline--traffic element graph generation is evaluated on both subsets.

\begin{dualtopoalgorithmfloat}
\noindent\begin{minipage}{\linewidth}
\begin{nonfloatalgorithm}{DualTopo: Lane Segment Graph Prediction}
\label{alg:dualtopo}
\begin{algorithmic}[1]
\REQUIRE Scene sequences $\{\mathcal{S}_1,\ldots\}$; action latent $\Psi$; mode $\in\{\text{train},\text{inference}\}$
\ENSURE Lane segment graph $\mathcal{G}^T = (\mathcal{V}, \mathcal{E})$ for each frame
\FOR{each sequence $\mathcal{S}$, frame $T$ in order}
  \STATE $\mathbf{F}^T \leftarrow \text{PV-to-BEV Encoder}(\mathcal{I}^T)$
  \STATE $\mathbf{Q}_1^T \leftarrow \text{TransLayer}_0(\mathbf{Q}_0, \mathbf{F}^T)$ \COMMENT{init queries}
  \IF{$T$ is first frame}
    \STATE $\tilde{\mathbf{Q}}_1^T \leftarrow \mathbf{Q}_1^T$, $\tilde{\mathbf{F}}^T \leftarrow \mathbf{F}^T$ \COMMENT{no replacement}
  \ELSE
    \STATE $\tilde{\mathbf{Q}}^T, \tilde{\mathbf{F}}^T \leftarrow \text{QWM, BWM}(\tilde{\mathbf{Q}}_L^{T-1}, \tilde{\mathbf{F}}^{T-1}; \Psi)$
    \STATE $\tilde{\mathbf{Q}}^T \leftarrow \tilde{\mathbf{Q}}^T + \lambda\,\mathcal{F}_g(\tilde{\mathbf{Q}}^T, \mathbf{A}^{T-1})$
      \STATE $\tilde{\mathbf{F}}^T \leftarrow \text{GRU}(\tilde{\mathbf{F}}^T, \mathbf{F}^T)$
    \STATE Replace the lowest-conf. $K$ queries of $\mathbf{Q}_1^T$ with the $K$ highest-conf. queries of $\tilde{\mathbf{Q}}^T$, yielding $\tilde{\mathbf{Q}}_1^T$
  \ENDIF
  \FOR{$i = 1$ \TO $L-1$}
    \STATE $\tilde{\mathbf{Q}}_{i+1}^T \leftarrow \text{TransLayer}_i(\tilde{\mathbf{Q}}_i^T, \tilde{\mathbf{F}}^T)$
    \IF{mode is train}
      \STATE $\mathbf{Q}_{i+1}^T \leftarrow \text{TransLayer}_i(\mathbf{Q}_i^T, \mathbf{F}^T)$ \COMMENT{shared weights}
    \ENDIF
  \ENDFOR
  \STATE $\mathcal{G}^T,\mathbf{A}^T \leftarrow \text{Heads}(\tilde{\mathbf{Q}}_L^T)$
  \IF{mode is train}
    \STATE Update $\alpha_1\mathcal{L}_{ls}(\tilde{\mathbf{Q}}_L^T) + \mathcal{L}_{ls}(\mathbf{Q}_L^T) + \mathbb{I}[T>1]\alpha_2\mathcal{L}_{\text{latent}}$
  \ENDIF
  \STATE Cache $\tilde{\mathbf{Q}}_L^T$, $\tilde{\mathbf{F}}^T$, and predicted $\mathbf{A}^T$ for the next frame
  \STATE \textbf{Output} $\mathcal{G}^T$ if mode is inference
\ENDFOR
\end{algorithmic}
\end{nonfloatalgorithm}
\end{minipage}

\end{dualtopoalgorithmfloat}

\textbf{Tasks and Metrics.} We follow the OpenLane-V2 v2.1.0 evaluation protocol \cite{wang2024openlane}.

\emph{Lane segment graphs.} Following LaneSegNet \cite{li2024lanesegnet}, we report lane segment AP$_{ls}$, pedestrian crossing AP$_{ped}$, and their mean (mAP). AP$_{ls}$ averages AP at distance thresholds of 1, 2, and 3\,m. For a predicted and ground-truth segment pair, let $\hat{\mathbf P}_s$ and $\mathbf P_s$ be their point sequences for the centerline ($s=c$), left boundary ($s=l$), or right boundary ($s=r$). The centerline term is $d_c=D_{\mathrm{F}}(\hat{\mathbf P}_c,\mathbf P_c)$, where $D_{\mathrm{F}}$ is the discrete Fr\'echet distance. The boundary terms are $d_l=D_{\mathrm{C}}(\hat{\mathbf P}_l,\mathbf P_l)$ and $d_r=D_{\mathrm{C}}(\hat{\mathbf P}_r,\mathbf P_r)$, where $D_{\mathrm{C}}$ is the Chamfer distance. The lane segment distance is $D_{ls}=\tfrac{1}{2}(d_c+d_l+d_r)$. AP$_{ped}$ averages AP at Chamfer distance thresholds of 0.5, 1.0, and 1.5\,m. TOP$_{lsls}$ measures directed connectivity between lane segments using the detection matches.

\emph{Centerline and traffic element graphs.} As in TopoNet \cite{li2023graph}, DET$_l$ averages centerline AP over discrete Fr\'echet distance thresholds of 1, 2, and 3\,m. DET$_t$ averages traffic element AP over attributes, using a bounding box distance of $1-\mathrm{IoU}$ with a threshold of 0.75. TOP$_{ll}$ scores directed links between centerlines, and TOP$_{lt}$ scores bipartite links between centerlines and traffic elements.

\emph{Topology and overall score.} After matching predicted vertices to ground truth, the evaluator ranks each vertex's candidate links by confidence and averages the resulting APs across vertices. Directed topology evaluates both incoming and outgoing edges. The OpenLane-V2 Score (OLS) combines centerline and traffic element detection with the square roots of the corresponding topology scores:
\begin{equation}
\label{eq:ols}
\mathrm{OLS}=\frac{1}{4}\Bigl[\sqrt{\mathrm{TOP}_{ll}}+\sqrt{\mathrm{TOP}_{lt}}+\mathrm{DET}_{l}+\mathrm{DET}_{t}\Bigr].
\end{equation}

\subsection{Implementation Details}\label{sec:implementation}
Our PV-to-BEV encoder combines a pre-trained ResNet-50 \cite{he2016deep}, a feature pyramid network (FPN) \cite{lin2017feature}, and BEVFormer \cite{li2022bevformer}. It produces a BEV grid with $(H,W,C)=(200,100,256)$, covering $\pm50$\,m longitudinally and $\pm25$\,m laterally. The decoder has $L=6$ layers and uses lane attention \cite{li2024lanesegnet} within a Deformable DETR architecture. We use $N\!=\!200$ query slots and propagate the $K\!=\!66$ highest-confidence historical queries, corresponding to 33\% of all slots. The QWM and BWM each contain two Transformer layers. We set $\lambda=0.25$ for directed topology propagation. The BWM applies average pooling to reduce the grid to $100\!\times\!50$, then restores $200\!\times\!100$ resolution by bilinear interpolation. The LaneSegNet implementation on SubsetA uses $M=10$ 3D points per centerline and boundary; centerline experiments on SubsetB use eleven 2D points per centerline.

Training follows a two-stage schedule with $E=24$ total epochs. Stage~1 trains the init pathway on single frames for $E_w=12$ epochs. Stage~2 starts from those weights and jointly trains both pathways on streaming inputs for another 12 epochs. Training uses eight NVIDIA V100 GPUs. We use AdamW \cite{diederik2015adam} with an initial learning rate of $2\!\times\!10^{-4}$ and cosine annealing. The weights $\alpha_1$ and $\alpha_2$ are 1.0 and 0.3, respectively.

\subsubsection{Streaming Training and Inference}

Algorithm~\ref{alg:dualtopo} summarizes DualTopo for streaming lane segment graph prediction during training and inference. Training uses this procedure in Stage~2, covering epochs $E_w{+}1{\ldots}E$.

\subsection{Main Results}\label{sec:main_results}

\textbf{Lane Segment Graph Generation (SubsetA).} DualTopo improves lane detection and connectivity through both output pathways (Table~\ref{tab:comparison_lane_segment}). The temporal pathway achieves 40.5 mAP and 32.1 TOP$_{lsls}$, exceeding the strongest competing scores by 3.9 and 1.3 percentage points. DualTopo also improves single-frame prediction: its init pathway raises mAP from 32.6 to 36.9 and TOP$_{lsls}$ from 25.4 to 28.4 over LaneSegNet, yielding gains of 4.3 and 3.0 percentage points, respectively. Joint training thus strengthens current-frame prediction while improving the primary temporal output.

\noindent\textbf{Centerline and Traffic Element Graph Generation.} DualTopo transfers to four centerline backbones, with gains in lane detection, lane-to-lane topology, and OLS on both subsets (Table~\ref{tab:centerline_full_comparison}). On SubsetA, DET$_l$ gains range from 1.7 to 5.6 percentage points, and TOP$_{ll}$ gains range from 0.8 to 8.0 points. OLS improves by 1.1--4.3 points, reaching 50.7 with TopoPoint. On SubsetB, DET$_l$ gains range from 3.1 to 8.4 percentage points, and TOP$_{ll}$ gains range from 5.6 to 10.3 points. OLS improves by 3.0--5.7 points. RATopo-based DualTopo achieves the strongest overall performance, with 55.0 OLS, 41.5 DET$_l$, and 45.9 TOP$_{ll}$. These results demonstrate that DualTopo readily adapts to a range of mainstream lane topology reasoning methods and delivers substantial improvements across both subsets. 

\subsection{Ablation Studies}
We evaluate module combinations (Table~\ref{tab:Ablation_modules}), transferability (Table~\ref{tab:streammapnet_generality}), and training and model configurations (Table~\ref{tab:world_model_config}).

\noindent\textbf{Module Combinations and Parallel Supervision.} Table~\ref{tab:Ablation_modules} shows that PQS and the world models improve both output pathways. The second row enables only Tem., using BEV warping for temporal propagation. This configuration reduces single-frame mAP and TOP$_{lsls}$ from 32.6 and 25.4 to 29.8 and 23.2. Adding PQS raises these scores to 36.4 and 27.8, while temporal prediction reaches 38.6 mAP and 30.5 TOP$_{lsls}$. Adding both world models to PQS raises temporal mAP and TOP$_{lsls}$ from 38.6/30.5 to 40.5/32.1, gains of 1.9/1.6 points. The complete model gains 6.5/4.0 points over temporal-only training. The init pathway simultaneously reaches 36.9 mAP and 28.4 TOP$_{lsls}$, improving on the single-frame baseline by 4.3 and 3.0 points. With both world models present, adding PQS increases init-path mAP from 34.2 to 36.9, confirming the value of independent query supervision alongside latent propagation.

\noindent\textbf{Generality on StreamMapNet.} Table~\ref{tab:streammapnet_generality} evaluates transfer to StreamMapNet \cite{yuan2024streammapnet} on nuScenes. Dual world models increase AP by 2.5 and 2.3 points on the old and new splits, respectively. Adding PQS increases the total gains to 4.2 and 3.5 points.

\noindent\textbf{Training Schedule and Action Conditioning.} Table~\ref{tab:world_model_config}(a) evaluates the two-stage schedule and relative ego-motion conditioning. Compared with 24 epochs of streaming training from scratch, the 12+12-epoch schedule improves temporal mAP and TOP$_{lsls}$ by 2.8 and 1.9 points. Relative-pose conditioning improves mAP from 38.1 to 40.5 and TOP$_{lsls}$ from 30.6 to 32.1. Single-frame pre-training establishes reliable initialized queries, while ego-motion conditions the learned temporal transition.

\noindent\textbf{Training Strategies.} Table~\ref{tab:world_model_config}(b) compares parallel training with randomly alternating the two pathways. Parallel supervision improves temporal mAP and TOP$_{lsls}$ by 2.2 and 1.0 points, respectively. It also raises single-frame mAP from 35.5 to 36.9 and TOP$_{lsls}$ from 27.8 to 28.4. Thus, joint optimization strengthens both pathways within the shared decoder.

\noindent\textbf{World Model Architecture.} Table~\ref{tab:world_model_config}(c) compares linear, MLP, and Transformer world models. The Transformer achieves the highest scores, improving temporal mAP and TOP$_{lsls}$ over linear prediction by 2.3 and 2.1 points. This supports attention-based modeling of the latent state transition.

\noindent\textbf{BEV History Length.} Table~\ref{tab:world_model_config}(d) evaluates one, two, four, and eight history frames. Two frames yield the best tested scores of 40.8 mAP and 32.4 TOP$_{lsls}$, improving both metrics by 0.3 points over the one-frame default. A short history supplies effective temporal context; we retain one frame in the default configuration.

\noindent\textbf{World Model Depth.} Table~\ref{tab:world_model_config}(e) shows that deeper world models can further improve prediction accuracy. We use two layers as the default.

\noindent\textbf{Temporal Query Count.} Table~\ref{tab:world_model_config}(f) varies the number of propagated queries with $N=200$ total slots. The default $K=66$ achieves the highest tested mAP and TOP$_{lsls}$, at 40.5 and 32.1.

\subsection{Temporal Consistency}\label{sec:temporal_consistency}
After ego-motion alignment, \emph{Jitter} measures symmetric mean nearest-point centerline distance in metres within the common visible region. It is averaged over the lane pairs successfully matched to the same GT lane in both frames for each model. \emph{Conditional warp} reports the percentage of these pairs with displacement at most 1\,m. A shared external tracker assigns scene-local identities by Hungarian association of ego-aligned geometry. Inspired by C-mAP in MapTracker~\cite{chen2024maptracker}, \emph{Temporal DET$_l$} measures detection AP while requiring each GT lane to retain its initially associated prediction identity. \emph{Temporal TOP$_{ll}$} measures incoming/outgoing adjacency AP for unchanged GT relations, using the lower edge confidence across the two frames. \emph{Persistent edges} and \emph{persistent paths} measure recovery of GT edges and reachable node pairs present in both frames. A relation counts as persistent when both predicted graphs recover it.

DualTopo improves temporal geometry and connectivity across the three evaluated backbones (Figure~\ref{fig:three_backbone_temporal}). Jitter decreases by 0.041--0.110\,m, while persistent-path recall increases by 2.7--8.8 percentage points. Temporal detection and topology scores also improve.

\subsection{Qualitative Results}

\noindent\textbf{BEV propagation.} Figure~\ref{fig:wm_warping_bev} connects propagated BEV features with the resulting lane graphs. Coordinate warping leaves unsupported boundary regions, whereas the world model predicts features across the current grid. In the four examples, DualTopo recovers more complete lane geometry and connections missing from the Warping predictions.

\noindent\textbf{Robustness to occlusion.} Figure~\ref{fig:vis} visualizes how temporal information improves robustness in occluded regions. By exploiting historical observations, DualTopo preserves more complete lane geometry and connectivity where current-frame views are obstructed. LaneSegNet misses lane segments in the left example and introduces a spurious branch absent from GT in the right example, whereas DualTopo recovers lane structure that more closely matches the ground truth.

\begin{figure}[!t]
% OLS: Table II SubsetB. FPS: archived author-provided A800 results; one decimal.
\centering
\captionof{table}{Accuracy and inference speed on OpenLane-V2 SubsetB. FPS is measured on an NVIDIA A800. Gray rows denote DualTopo.}\label{tab:efficiency_comparison}
\small\setlength{\tabcolsep}{6pt}\renewcommand{\arraystretch}{1.08}
\begin{tabular}{@{}llrr@{}}
\toprule
Backbone & Variant & OLS$\uparrow$ & FPS$\uparrow$ \\ \midrule
\multirow{2}{*}{RATopo} & Baseline & 50.3 & \textbf{5.6} \\
& \cellcolor{gray!8}\textbf{DualTopo} & \cellcolor{gray!8}\textbf{55.0} & \cellcolor{gray!8}4.8 \\
\addlinespace[4pt]
\multirow{2}{*}{TopoLogic} & Baseline & 42.3 & \textbf{13.2} \\
& \cellcolor{gray!8}\textbf{DualTopo} & \cellcolor{gray!8}\textbf{48.0} & \cellcolor{gray!8}11.2 \\
\addlinespace[4pt]
\multirow{2}{*}{TopoPoint} & Baseline & 49.2 & \textbf{9.1} \\
& \cellcolor{gray!8}\textbf{DualTopo} & \cellcolor{gray!8}\textbf{52.2} & \cellcolor{gray!8}7.8 \\
\addlinespace[4pt]
\multirow{2}{*}{TopoFG} & Baseline & 45.4 & \textbf{6.2} \\
& \cellcolor{gray!8}\textbf{DualTopo} & \cellcolor{gray!8}\textbf{48.8} & \cellcolor{gray!8}5.4 \\
\bottomrule
\end{tabular}

\end{figure}

\noindent\textbf{Sequence consistency.} Figure~\ref{fig:stream_compare} compares predictions over five consecutive frames in two scenes. DualTopo maintains more complete lane segments and more consistent junction connectivity as the ego vehicle moves. LaneSegNet exhibits intermittent missed detections and fragmented connections in the highlighted regions.

\subsection{Accuracy and Inference Speed}\label{sec:efficiency}

Table~\ref{tab:efficiency_comparison} pairs the SubsetB OLS scores from Table~\ref{tab:centerline_full_comparison} with batch-one throughput from image encoding through the final prediction head on an NVIDIA A800. Across the four backbones, DualTopo improves OLS by 3.0--5.7 points while retaining 84.8--87.4\% of baseline throughput.

\section{Conclusion}
We presented DualTopo, a framework for spatio-temporal lane graph generation that combines dual latent world models with parallel query supervision. Motion-conditioned query and BEV prediction carries scene geometry and directed topology across frames, while independent supervision within a shared decoder strengthens both single-frame and temporal prediction. Experiments on OpenLane-V2 demonstrate improved graph generation and temporal consistency. Adaptation to multiple centerline backbones further establishes the generality of the framework.

\bibliographystyle{IEEEtran}
\bibliography{main}
\clearpage
\section*{Supplementary Material}
\appendix

\subsection{Empirical Validation: Temporal Queries Dominate Hungarian Matching}
\label{sec:matching_analysis}

\noindent\textbf{Setup.} We train a standard stream-based baseline (i.e., the \textit{Temporal}-only row in Table~\ref{tab:Ablation_modules} of the main paper, without the proposed parallel query supervision) using $K\!=\!66$ temporal queries and $N\!-\!K\!=\!134$ init queries ($N\!=\!200$ total). Per-frame matching statistics are collected from the last decoder layer during Stage~2 streaming training across 4 GPUs. We exclude first frames (which have no temporal context) and analyze 21,643 non-first-frame records spanning 696 scenes.

\begin{figure}[t]
    \centering
    \includegraphics[width=0.55\columnwidth]{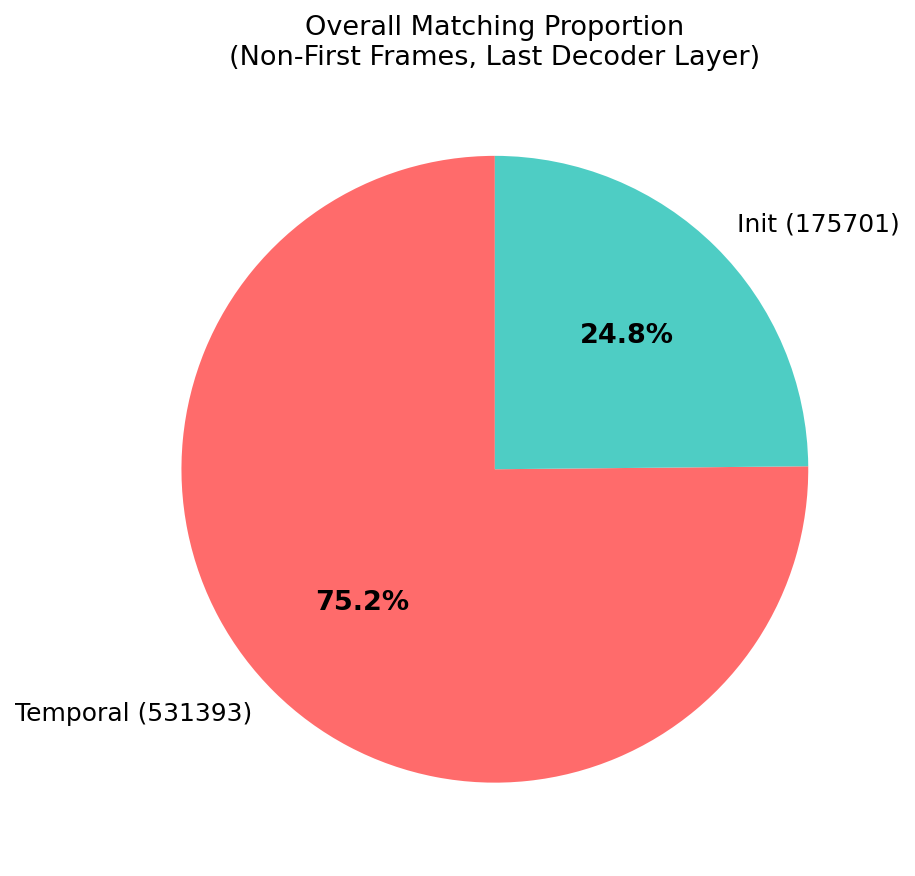}
    \caption{Overall Hungarian matching proportion during stream-based training. Although temporal queries comprise only 33\% of all queries, they account for 75.2\% of all GT matches, leaving init queries severely under-supervised.}
    \label{fig:matching_pie}
\end{figure}

\noindent\textbf{Overall Matching Proportion.}
Figure~\ref{fig:matching_pie} shows that temporal queries, despite comprising only 33\% of all queries, account for \textbf{75.2\%} of all GT matches (531,393 out of 707,094 total). These counts quantify the allocation of matched supervision between the query groups. The module-combination experiment separately measures its effect on prediction accuracy.

\begin{figure*}[!t]
    \centering
    \includegraphics[width=0.63\textwidth]{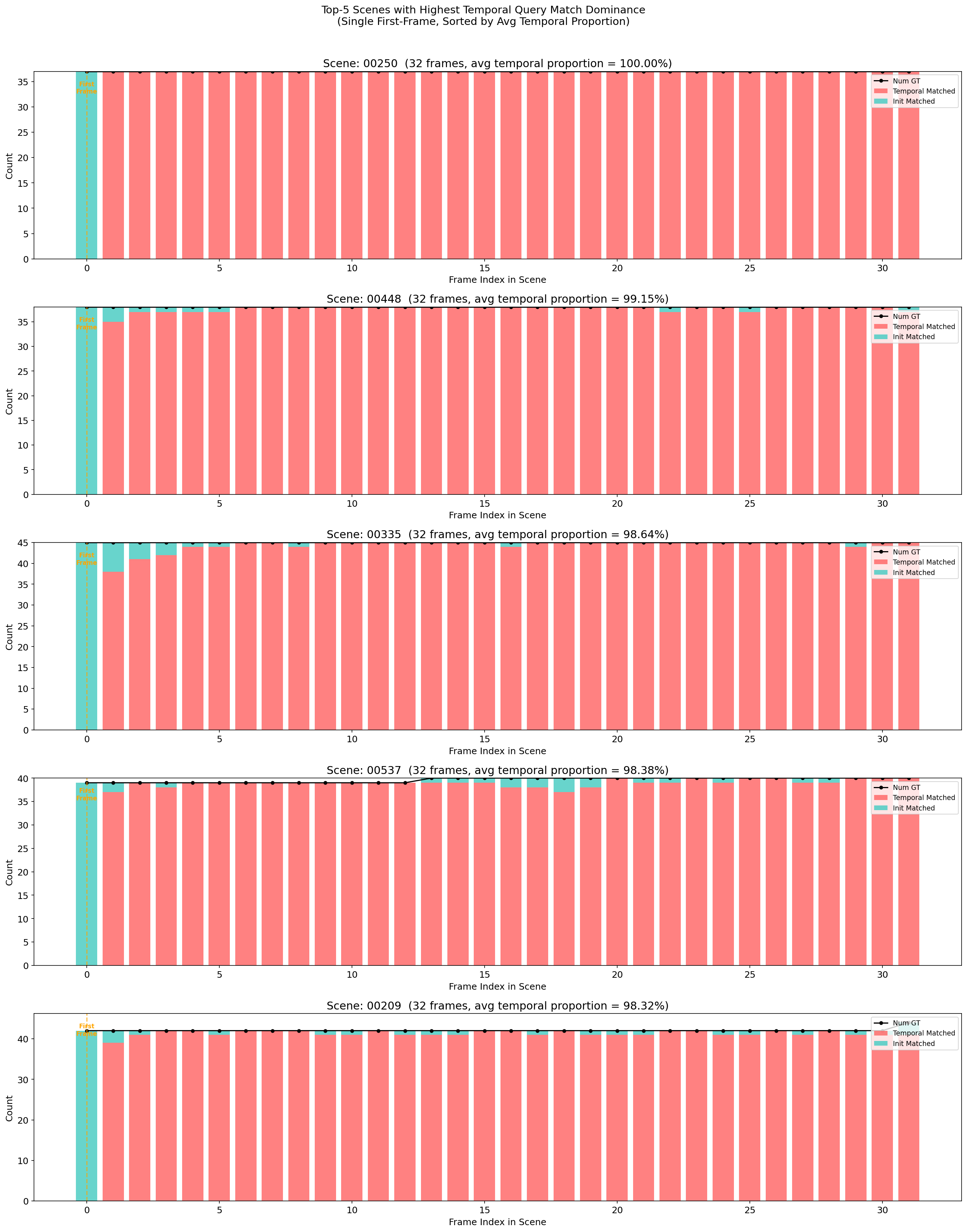}
    \caption{Frame-by-frame matching breakdown for the top-5 temporal-dominant scenes. Red bars: GT matched to temporal queries; teal bars: GT matched to init queries; black line: number of GT lane segments. After the first frame (orange dashed line), temporal queries dominate matching in these selected extreme cases.}
    \label{fig:matching_scenes}
\end{figure*}

\noindent\textbf{Per-Scene Matching Dominance.}
Figure~\ref{fig:matching_scenes} shows the five scenes with the highest temporal match share. In these selected extreme cases, temporal queries receive over 97\% of GT matches after the first frame. The pooled 75.2\% statistic above summarizes all analyzed non-first-frame records.

\noindent\textbf{Implication for Dual-Pathway Design.} Separate Hungarian assignments provide initialized queries with direct supervision. In Table~\ref{tab:Ablation_modules}, adding PQS to ordinary temporal propagation increases init-path mAP from 29.8 to 36.4. The matching statistics and model-level comparisons jointly motivate separate supervision for the initialized pathway.

\subsection{Action Latent}

Action-conditioning and depth results are reported in Table~\ref{tab:world_model_config}(a) and (e) of the main manuscript. Relative pose conditioning improves the default model, whereas depth changes produce mixed detection and topology outcomes. The following trajectory condition uses future ego-vehicle waypoints.

\noindent\textbf{Relative Pose.} The relative pose is defined as the rigid transformation from the previous frame's coordinate system to the current frame's coordinate system:
\begin{equation}
\begin{aligned}
\mathbf{T}_{\text{prev}}^{\text{global}}
&= \begin{bmatrix}
\mathbf{R}_{\text{prev}} & \mathbf{t}_{\text{prev}} \\
\mathbf{0} & 1
\end{bmatrix}, \\
\mathbf{T}_{\text{global}}^{\text{curr}}
&= \begin{bmatrix}
\mathbf{R}_{\text{curr}}^\top & -\mathbf{R}_{\text{curr}}^\top \mathbf{t}_{\text{curr}} \\
\mathbf{0} & 1
\end{bmatrix}.
\end{aligned}
\end{equation}
\begin{equation}
\begin{aligned}
\mathbf{T}_{\text{prev}}^{\text{curr}}
&= \mathbf{T}_{\text{global}}^{\text{curr}} \mathbf{T}_{\text{prev}}^{\text{global}}, \\
\Psi &= \text{Flatten}\!\left(
\left[\mathbf{R}_{\text{prev}}^{\text{curr}} \mid
\mathbf{t}_{\text{prev}}^{\text{curr}}\right]\right) \in \mathbb{R}^{12}.
\end{aligned}
\end{equation}
where $\mathbf{R}_{\text{prev}}^{\text{curr}}$ and $\mathbf{t}_{\text{prev}}^{\text{curr}}$ are the rotation and translation components of $\mathbf{T}_{\text{prev}}^{\text{curr}}$, respectively. The resulting 12-dimensional vector $\Psi$, obtained by flattening the first three rows of the homogeneous transformation matrix, is used as the action latent for both the QWM and BWM.

\noindent\textbf{Trajectory.} As an alternative, we explored using future ego-vehicle positions as the action condition, following prior work \cite{li2024enhancinge2e}. We extract positions over a 3-second horizon (6 frames) and transform them into the current BEV coordinate system to produce waypoints $\mathbf{W}_T=\{\mathbf{w}_T^1,\ldots,\mathbf{w}_T^{N_w}\}$ ($N_w=6$). For the final five frames of a scene where future positions are unavailable, we linearly extrapolate the missing waypoints using the last observed displacement. A trajectory query is initialized randomly and refined via cross-attention with BEV features and lane segment queries; an MLP then decodes the trajectory. The trajectory prediction loss is:
\begin{equation}
\mathcal{L}_{\text{traj}} = \frac{1}{N_w} \sum_{i=1}^{N_w} \left\| \mathbf{w}_T^i - \hat{\mathbf{w}}_T^i \right\|_1
\end{equation}
The predicted waypoints are concatenated with queries and BEV features along the channel dimension before being fed to the world models.

\begin{figure*}[!t]
    \centering
    \includegraphics[width=0.81\textwidth]{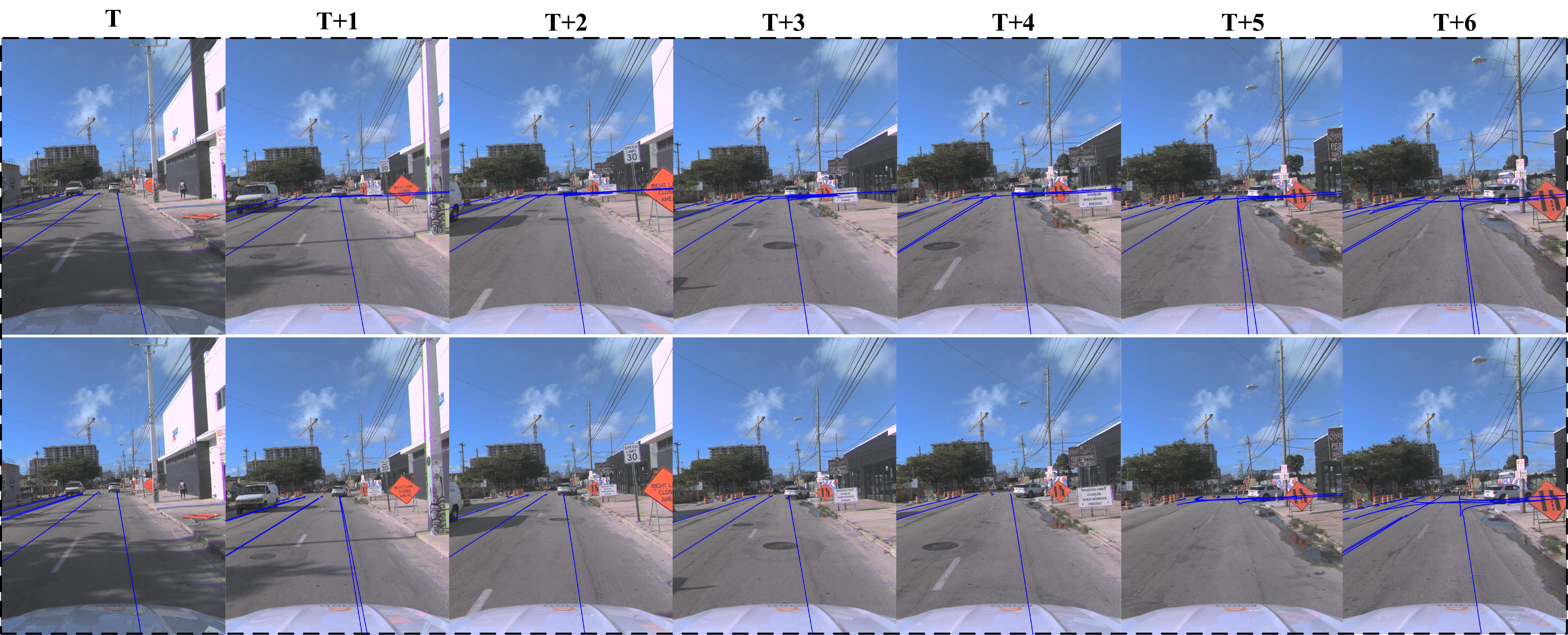}
    \caption{Front-view visualization of DualTopo (top) and the stream-based baseline (bottom) across sequential frames. DualTopo detects intersection lane structures earlier and maintains temporally consistent graph predictions, while the baseline exhibits delayed detection and poor temporal consistency.}
    \label{fig:front_vis}
    \par\vspace{8pt}
    \centering
    \includegraphics[width=0.73\textwidth]{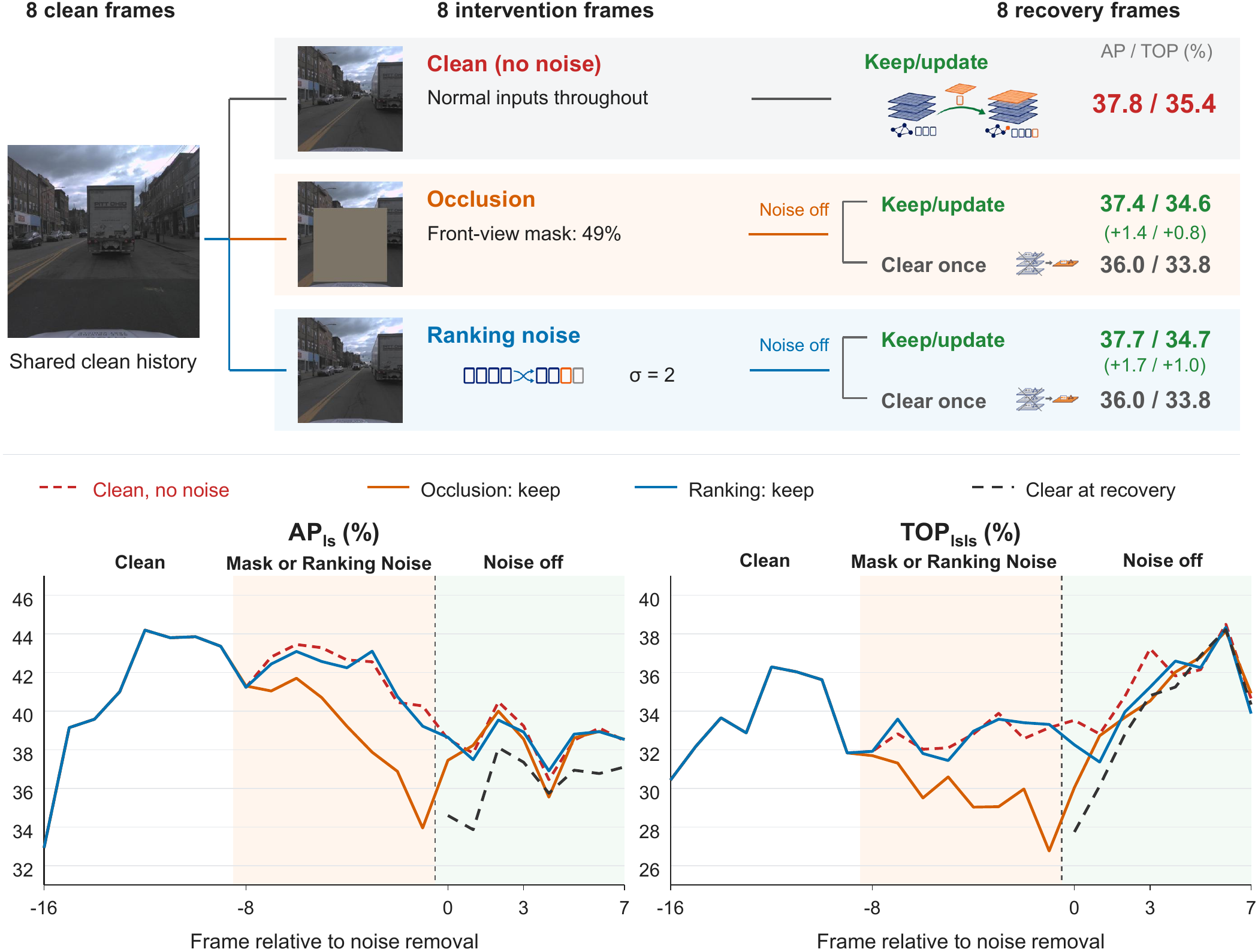}
    \caption{Robustness to confidence noise and occlusion. Confidence-based memory-ranking noise has little effect on temporal prediction, while front-camera occlusion causes a temporary reduction. After either perturbation is removed, continued memory updates rapidly restore near-clean performance. Red marks the clean control; orange and blue retain and update memory after noise removal. Dark dashed curves clear memory once at recovery onset, then resume updates; the two clear-memory curves coincide. Boxes report recovery-window AP$_{ls}$/TOP$_{lsls}$ (\%), with green gains relative to clearing. Curves pool ten scenes at each frame.}
    \label{fig:noise_recovery}
\end{figure*}

\subsection{BEV World Model: Resolution Downsampling}

To reduce GPU memory overhead during latent state prediction, the BWM operates on a downsampled BEV representation. Specifically, the action-aware historical BEV memory $\hat{\mathbf{F}}_{\text{mem}}^{T-1} \in \mathbb{R}^{H \times W \times C}$ (resolution $200\!\times\!100$) is downsampled to $100\!\times\!50$ via average pooling before being passed to the BWM:
\begin{equation}
\begin{aligned}
\hat{\mathbf{F}}_{\text{mem, down}}^{T-1}
&= \text{AvgPool}(\hat{\mathbf{F}}_{\text{mem}}^{T-1}), \\
\tilde{\mathbf{F}}_{\text{bev, down}}^{T}
&= \textbf{BWM}(\hat{\mathbf{F}}_{\text{mem, down}}^{T-1}).
\end{aligned}
\end{equation}
The predicted temporal BEV features are then restored to the original $200\!\times\!100$ resolution via bilinear interpolation:
\begin{equation}
\tilde{\mathbf{F}}_{\text{bev}}^{T} = \text{Interp}(\tilde{\mathbf{F}}_{\text{bev, down}}^{T})
\end{equation}
The supervision loss $\mathcal{L}_{\text{bev}}=\operatorname{MSE}(\tilde{\mathbf{F}}_{\text{bev}}^T,\mathbf{F}_{\text{bev}}^T)$ is computed at the full $200\!\times\!100$ resolution and averaged over all spatial positions and feature channels. This design reduces the BWM's computational cost while preserving the spatial fidelity of the predicted temporal BEV features.

\subsection{GRU-Based BEV Feature Fusion}

After obtaining temporal BEV features $\tilde{\mathbf{F}}_{\text{bev}}^T$ from the BWM and current-frame BEV features $\mathbf{F}_{\text{bev}}^T$ from the PV-to-BEV encoder, we fuse them via a gated recurrent unit (GRU) \cite{chung2014empirical}:
\begin{equation}
\mathbf{F}_{\text{fused}}^T = \text{GRU}(\tilde{\mathbf{F}}_{\text{bev}}^T,\, \mathbf{F}_{\text{bev}}^T)
\end{equation}
The GRU takes the temporal BEV features $\tilde{\mathbf{F}}_{\text{bev}}^T$ (carrying historical temporal context propagated by the BWM) as the recurrent input and the current-frame BEV features $\mathbf{F}_{\text{bev}}^T$ as the state context, adaptively gating their contributions and producing the fused BEV representation $\mathbf{F}_{\text{fused}}^T$ used as the BEV context for the temporal decoder pathway.

\subsection{Transformation Loss}

\noindent\textbf{Lane segment graphs.} Transformation losses supervise the QWM by requiring temporal queries $\tilde{\mathbf{Q}}^T$ to correctly predict lane segment attributes after temporal transformation. We apply MLPs to decode lane segment coordinates, classes, boundary types, and BEV semantic masks from $\tilde{\mathbf{Q}}^T$:
{\setlength{\jot}{0pt}\setlength{\abovedisplayskip}{4pt}\setlength{\belowdisplayskip}{4pt}\setlength{\abovedisplayshortskip}{2pt}\setlength{\belowdisplayshortskip}{4pt}
\begin{equation}
    \begin{split}
    \tilde{\mathbf{L}}^c_T
    &= \text{Denorm}\!\left(\text{Sigmoid}\!\left(
       \text{MLP}_{\text{reg}}(\tilde{\mathbf{Q}}^T)\right.\right.\\[-2pt]
    &\qquad\left.\left.{}+\text{InvSigmoid}(\mathbf{R}^S_C)\right)\right)\\
    \tilde{\mathbf{L}}^l_T &= \tilde{\mathbf{L}}^c_T + \text{MLP}_{\text{offset}}(\tilde{\mathbf{Q}}^T)\\
    \tilde{\mathbf{L}}^r_T &= \tilde{\mathbf{L}}^c_T - \text{MLP}_{\text{offset}}(\tilde{\mathbf{Q}}^T)\\
    \tilde{\mathbf{L}}_T &= \text{Concat}(\tilde{\mathbf{L}}^c_T, \tilde{\mathbf{L}}^l_T, \tilde{\mathbf{L}}^r_T)\\
    \tilde{\mathit{Class}}_T &= \text{MLP}_{\text{cls}}(\tilde{\mathbf{Q}}^T), \quad \tilde{\mathit{Type}}_T = \text{MLP}_{\text{bcls}}(\tilde{\mathbf{Q}}^T)\\
    \tilde{\mathbf{M}}_T &= \text{Sigmoid}\!\left(\text{MLP}_{\text{mask}}(\tilde{\mathbf{Q}}^T) \otimes \tilde{\mathbf{F}}_{\text{bev}}^T\right)
    \end{split}
\end{equation}
}
where $\mathbf{R}^S_C$ denotes centerline reference points for temporal queries, and InvSigmoid is the inverse sigmoid function. The transformation losses are:
\begin{equation}
    \begin{split}
	\mathcal{L}_{\text{coord}} &= \mathcal{L}_{L1}(\tilde{\mathbf{L}}_T,\, \mathbf{L}_T)\\
	\mathcal{L}_{\text{cls}} &=  \kappa_1\mathcal{L}_{\text{Focal}}(\tilde{\mathit{Class}}_T, \mathit{Class}_T)\\
    &\qquad{}+\kappa_2\mathcal{L}_{\text{CE}}(\tilde{\mathit{Type}}_T, \mathit{Type}_T)\\
	\mathcal{L}_{\text{mask}} &= \kappa_3\mathcal{L}_{\text{CE}}(\tilde{\mathbf{M}}_T, \mathbf{M}_T)\\
    &\qquad{}+\kappa_4\mathcal{L}_{\text{Dice}}(\tilde{\mathbf{M}}_T, \mathbf{M}_T)\\
	\mathcal{L}_{\text{query}} &= \kappa_5\mathcal{L}_{\text{coord}} + \kappa_6\mathcal{L}_{\text{cls}} + \kappa_7\mathcal{L}_{\text{mask}}
    \end{split}
\end{equation}
where $\kappa_1\!=\!1.5$, $\kappa_2\!=\!0.1$, $\kappa_3\!=\!1.0$, $\kappa_4\!=\!1.0$, $\kappa_5\!=\!0.025$, $\kappa_6\!=\!1.0$, $\kappa_7\!=\!3.0$.

\noindent\textbf{Centerline and traffic element graphs.} The backbone's final-layer lane heads project propagated queries to class logits $\tilde{\mathbf c}^{T}$ and centerline coordinates $\tilde{\mathbf P}^{T}$ using the transported references. Their transformation supervision is
\begin{equation}
\begin{aligned}
\mathcal L_{\mathrm{trans}}^{\mathrm{cl}}
&=\alpha_2\bigl[\mathcal L_{\mathrm{cls}}^{\mathrm{base}}(\tilde{\mathbf c}^{T},\mathbf y^{T})\\
&\qquad+\mathcal L_{\mathrm{reg}}^{\mathrm{base}}(\tilde{\mathbf P}^{T},\mathbf P_{\mathrm{tar}}^{T})\bigr],
\end{aligned}
\end{equation}
where $\alpha_2=0.3$, $\mathbf y^{T}$ and $\mathbf P_{\mathrm{tar}}^{T}$ denote valid classification and geometry targets. The baseline loss functions retain their original weights, coordinate normalization, and valid-query masks. A propagated query inherits its lane ID from the previous frame's lane assignment. For SubsetB, its target is the current-frame annotation of that lane when available. For a previously matched lane absent from the current annotations, RATopo, TopoPoint, and TopoFG use the previous ground-truth centerline transformed to the current ego frame, while TopoLogic masks the query from this loss. The decoder supervises LCLC and LCTE separately using the current-frame graph annotations.

\subsection{Prediction Heads}

\noindent\textbf{Lane segment graphs.} After the decoder, shared-weight MLPs decode lane segment attributes and topology from the updated queries $\mathbf{Q}$ (applicable to both the temporal pathway output $\tilde{\mathbf{Q}}^T_L$ and the init pathway output $\mathbf{Q}^T_L$):
\begin{equation}
    \begin{split}
    \mathbf{R}_c &= \text{Sigmoid}(\text{MLP}_{\text{pe}}(\mathbf{PE}))\\
    \tilde{\mathbf{L}}^c
    &= \text{Denorm}\!\left(\text{Sigmoid}\!\left(
       \text{MLP}_{\text{reg}}(\mathbf{Q})\right.\right.\\[-2pt]
    &\qquad\left.\left.{}+\text{InvSigmoid}(\mathbf{R}_c)\right)\right)\\
    \tilde{\mathbf{L}}^l &= \tilde{\mathbf{L}}^c + \text{MLP}_{\text{offset}}(\mathbf{Q})\\
    \tilde{\mathbf{L}}^r &= \tilde{\mathbf{L}}^c - \text{MLP}_{\text{offset}}(\mathbf{Q})\\
    \tilde{\mathbf{L}} &= \text{Concat}(\tilde{\mathbf{L}}^c, \tilde{\mathbf{L}}^l, \tilde{\mathbf{L}}^r)\\
    \tilde{\mathit{Class}} &= \text{MLP}_{\text{cls}}(\mathbf{Q}), \quad \tilde{\mathit{Type}} = \text{MLP}_{\text{bcls}}(\mathbf{Q})\\
    \tilde{\mathbf{M}} &= \text{Sigmoid}(\text{MLP}_{\text{mask}}(\mathbf{Q}) \otimes \mathbf{F}_{\text{bev}})\\
    \tilde{A}_{ij}
    &= \text{Sigmoid}\!\left(\text{MLP}_{\text{topo}}\!\left(
       \text{Concat}\!\left(\text{MLP}_{\text{pre}}(\mathbf{Q}_i),\right.\right.\right.\\[-2pt]
    &\qquad\left.\left.\left.\text{MLP}_{\text{suc}}(\mathbf{Q}_j)\right)\right)\right),
       \quad i,j=1,\ldots,N
    \end{split}
\end{equation}
where $\mathbf{R}_c$ denotes centerline reference points derived from the positional embedding $\mathbf{PE}$, and $\tilde{\mathbf{A}} \in [0,1]^{N \times N}$ is the predicted adjacency matrix encoding topological graph edges between lane segments with a confidence threshold of 0.5.

\noindent\textbf{Centerline and traffic element graphs.} For a decoded lane query $\mathbf q_i$ and current-frame traffic element query $\mathbf q_j^{\mathrm{te}}$, the shared baseline heads give
\begin{equation}
\begin{aligned}
\hat{\mathbf c}_i &= h_{\mathrm{cls}}(\mathbf q_i),\quad
\hat{\mathbf P}_i = h_{\mathrm{reg}}(\mathbf q_i,\mathbf r_i),\\
\hat A^{ll}_{ij} &= \operatorname{Sigmoid}\bigl(h_{ll}(\mathbf q_i,\mathbf q_j)\bigr),\\
\hat A^{lt}_{ij} &= \operatorname{Sigmoid}\bigl(h_{lt}(\mathbf q_i,\mathbf q_j^{\mathrm{te}})\bigr),
\end{aligned}
\end{equation}
where $\mathbf r_i$ is the lane reference. The functions $h$ denote each backbone's native prediction branches, including its feature projections and geometric inputs. The LCLC matrix $\hat{\mathbf A}^{ll}\in[0,1]^{N_l\times N_l}$ represents directed lane connectivity, and the LCTE matrix $\hat{\mathbf A}^{lt}\in[0,1]^{N_l\times N_t}$ represents lane--traffic element relations. Here $N_l$ and $N_t$ count lane and traffic element queries. The traffic element detector supplies boxes and class scores. Each pathway independently matches its lanes, while its LCTE loss also uses the detector's traffic element matches.

\subsection{Additional Qualitative Results}

Figure~\ref{fig:front_vis} shows that DualTopo detects intersection structures earlier and produces more temporally consistent graph predictions than the stream-based baseline.

\subsection{Robustness to Transient Noise}
DualTopo is robust to transient confidence perturbations and partial visual occlusion. We separately test front-camera occlusion covering 49\% of the image and confidence-based memory-ranking noise with Gaussian $\sigma=2$ on ten scenes. Each sequence contains eight clean frames, eight perturbed frames, and eight recovery frames. Figure~\ref{fig:noise_recovery} shows that ranking noise has little effect on AP$_{ls}$ and TOP$_{lsls}$, while occlusion causes a temporary reduction without catastrophic degradation. After noise removal, continued memory updates rapidly restore near-clean performance, even though recovery starts from a perturbed memory state. Compared with clearing memory once at recovery onset and then resuming updates, retaining it improves recovery-window AP$_{ls}$/TOP$_{lsls}$ by 1.4/0.8 points for occlusion and 1.7/1.0 points for ranking noise. These results show that useful history survives transient corruption and supports rapid recovery without a memory reset.

\end{document}